\documentclass{article}
\usepackage{nips14submit_e}
\usepackage{hyperref}
\usepackage{url}
\usepackage{times}
\usepackage{epsfig}
\usepackage{graphicx}
\usepackage{amsmath}
\usepackage{amssymb}
\usepackage{rotating}
\usepackage{caption}
\usepackage{subcaption}

\usepackage{xspace}
\makeatletter
\DeclareRobustCommand\onedot{\futurelet\@let@token\@onedot}
\def\@onedot{\ifx\@let@token.\else.\null\fi\xspace}

\def\eg{\emph{e.g}\onedot} 
\def\ie{\emph{i.e}\onedot}

\newcommand{\diag}[1]{\text{diag}\left(#1\right)}
\def\trans{\text{T}}
\makeatother

\title{Large Scale, Large Margin Classification using Indefinite Similarity Measures}

\author{Omid Aghazadeh and Stefan Carlsson}

\nipsfinalcopy 

\begin{document}

\maketitle

\begin{abstract}
Despite the success of the popular kernelized support vector machines, they have two major limitations:
they are restricted to Positive Semi-Definite (PSD) kernels, and their training complexity scales at least quadratically with the size of the data.
Many natural measures of similarity between pairs of samples are not PSD \eg invariant kernels, and those that are implicitly or explicitly defined by latent variable models.
In this paper, we investigate scalable approaches for using indefinite similarity measures in large margin frameworks.
In particular we show that a normalization of similarity to a subset of the data points constitutes a representation suitable for linear classifiers.
The result is a classifier which is competitive to kernelized SVM in terms of accuracy, despite having better training and test time complexities.
Experimental results demonstrate that on CIFAR-10 dataset, the model equipped with similarity measures invariant to rigid and non-rigid deformations, can be made more than 5 times sparser while being more accurate than kernelized SVM using RBF kernels.
\end{abstract}

\section{Introduction}
\label{paperE:sec:introduction}
Linear support vector machine (SVM) has become the classifier of choice for many large scale classification problems.
The main reasons for the success of linear SVM are its max margin property achieved through a convex optimization, a training time linear in the size of the training data, and a testing time independent of it.
Although the linear classifier operating on the input space is usually not very flexible, a linear classifier operating on a mapping of the data to a higher dimensional feature space can become arbitrarily complex.

Mixtures of linear classifiers has been proposed to increase the non-linearity of linear classifiers \cite{mixing_svm,AghazadehASC12}; which can be seen as feature mappings augmented with non-linear gating functions.
The training of these mixture models usually scales bilinearly with respect to the data and the number of mixtures.
The drawback is the non-convexity of the optimization procedures, and the need to know the (maximum) number of components beforehand.

Kernelized SVM maps the data to a possibly higher dimensional feature space, maintains the convexity, and can become arbitrarily flexible depending on the choice of the kernel function.
The use of kernels, however, is limiting.

Firstly, kernelized SVM has significantly higher training and test time complexities when compared to linear SVM. 
As the number of support vectors grows approximately linearly with the training data \cite{Steinwart04sparsenessof}, 
the training complexity becomes approximately somwehere between $\mathcal{O}(n^2)$ and $\mathcal{O}(n^3)$.
Testing time complexity scales linearly with the number of support vectors, 
bounded by $\mathcal{O}(n)$.

Secondly, the positive (semi) definite (PSD) kernels are sometimes not expressive enough to model various sources of variation in the data.
A recent study \cite{good_rec_nonmetric} argues that metric constraints are not necessarily optimal for recognition.
For example, in image classification problems, considering kernels as similarity measures, they cannot align exemplars, or model deformations when measuring similarities.
As a response to this, invariant kernels were introduced \cite{invariant_kernels} which are generally indefinite.
Indefinite similarity measures plugged in SVM solvers result in non-convex optimizations, unless explicitly made PSD, mainly using eigen decomposition methods \cite{sim_class}.
Alternatively, latent variable models have been proposed to address the alignment problem \eg \cite{FelzenszwalbGMR10,YangWVM12}.
In these cases, the dependency of the latent variables on the parameters of the model being learnt mainly has two drawbacks:
1) the optimization problem in such cases becomes 
non-convex, 
and 2) the cost of training becomes much higher than the case without the latent variables.

This paper aims to address these problems using explicit basis expansion.
We show that the resulting model:
1) has better training and test time complexities than kernelized SVM models, 
2) can make use of indefinite similarity measures without any need for removal of the negative eigenvalues, which requires the expensive eigen decomposition,
3) can make use of multiple similarity measures without losing convexity, and with a cost linear in the number of similarity measures.

Our contributions are: 1) proposing and analyzing Basis Expanding SVM (BE-SVM) regarding the aforementioned three properties, and 2) investigating the suitability of particular forms of invariant similarity measures for large scale visual recognition problems.

\section{Basis Expanding Support Vector Machine}
\label{paperE:sec:model}
\subsection{Background: SVM}
\label{paperE:sec:model:background}
Given a dataset $\mathcal{D}=\{(x_1,y_1), \ldots, (x_n,y_n)| x_i \in \mathcal{X},\, y_i \in \{-1,1\}\}$ the SVM based classifiers learn max margin binary classifiers.
The SVM classifier is $ f(x)=\langle w, x\rangle\geq 0$ \footnote{We omit the bias term for the sake of clarity.}.
The $w$ is learnt via minimizing $\frac{1}{2}\langle w,w \rangle + C\sum_i \ell_H(y_i, f(x))$, where $\ell_H(y,x) = \max(0,1-xy)$ is the Hinge loss.
Any positive semi definite (PSD) kernel $k:\mathcal{X}\times\mathcal{X}\rightarrow\mathbb{R}$ can be associated with a reproducing kernel hilbert space (RKHS) $\mathcal{H}$, and vice versa, that is 
$\langle \psi_\mathcal{H}(x),\psi_\mathcal{H}(y) \rangle = k(x,y)$, 
where $\psi_{\mathcal{H}}:\mathcal{X}\rightarrow \mathcal{H}$ is the implicitly defined feature mapping associated to $\mathcal{H}$ and consequently to $k(.,.)$.
Representer theorem states that in such a case, 
$\psi_\mathcal{H}(w)=\sum_i \gamma_i k(.,x_i)$ where $\gamma_i \in \mathbb{R}\,\, \forall i$.

For a particular case of $k(.,.)$, namely the linear kernel $k(\mathbf{x},\mathbf{y})=\mathbf{x}\cdot\mathbf{y}$ associated with an Euclidean space, linear SVM classifier is 
~$f_l(\mathbf{x})=\mathbf{w}^\trans\mathbf{x}\geq 0$
where $\mathbf{w}$ is given by minimizing the primal SVM objective:
~$\frac{1}{2}\|\mathbf{w}\|^2 + C\sum_i \ell_H(y_i,f_l(\mathbf{x}_i))$.

More generally, given an arbitrary PSD kernel $k(.,.)$, the kernelized SVM classifier is
~$ f_k(x) = \sum_i \alpha_i k(x,x_i) \geq 0 $
where $\alpha_i$s are learnt by minimizing the dual SVM objective: 
~$\frac{1}{2}\mathbf{\alpha}^\trans\mathbf{Y}\mathbf{K}\mathbf{Y}\mathbf{\alpha}- \|\mathbf{\alpha}\|_1,\, 0\leq \alpha_i \leq C, \, \mathbf{\alpha}^\trans\mathbf{y} = 0$
where $\mathbf{Y}=\diag{\mathbf{y}}$.

The need for positiveness of $k(.,.)$ is evident
in the dual SVM objective
where the quadratic regularizing term depends on the eigenvalues of $\mathbf{K}_{ij} = k(x_i,x_j)$. 
In case of indefinite $k(.,.)$s, the problem becomes non-convex and the inner products need to be re-defined, as there will be no associating RKHS to indefinite similarity measures.
Various workarounds for indefinite similarity measures exist, most of which involve expensive eigen decomposition of the gram matrix \cite{sim_class}.
A PSD kernel can be learnt from the similarity matrix, with some constraints \eg being close to the similarity matrix where closeness is usually measured by the Frobenius norm.
In case of Frobenius norm, the closed form solution is spectrum clipping, namely setting the negative eigenvalues of the gram matrix to 0 \cite{sim_class}. 
As pointed out in \cite{learning_kernels_from_indef}, there is no guarantee that the resulting PSD kernels are optimal for classification.
Nevertheless, jointly optimizing for a PSD kernel and the classifier \cite{learning_kernels_from_indef} is impractical for large scale scenarios.
We do not go into the details of possible re-formulations regarding indefinite similarity measures, but refer the reader to \cite{ong04leanpk,Indefinite_Kernels,sim_class} for more information.

Linear and Kernelized SVM have very different properties.
Linear SVM has a training cost of $\mathcal{O}(d_\mathbf{x}n)$ and a testing cost of $\mathcal{O}(d_\mathbf{x})$ where $d_\mathbf{x}$ is the dimensionality of $\mathbf{x}$ .
Kernelized SVM has a training complexity which is $\mathcal{O}(d_k (n n_{sv}) + n_{sv}^3)$ \cite{Keerthi06buildingsupport} where $d_k$ is the cost of evaluating the kernel for one pair of data, and $n_{sv}$ is the number of resulting support vectors.
The testing cost of kernelized SVM is $\mathcal{O}(d_kn_{sv})$.
Therefore, a significant body of research has been dedicated to reducing the training and test costs of kernelized SVMs by approximating the original problem. 

\subsection{Speeding up Kernelized SVM}
\label{paperE:sec:model:speedup}
A common approach for approximating the kernelized SVM problem is to restrict the feature mapping of $w$: 
$\psi_\mathcal{H}(w) \approx \psi_R(w) = \sum_{j=1}^J \beta_j \psi_\mathcal{H} (z_j)$ where $J < n$.
Methods in this direction either learn synthetic samples $z_j$ \cite{direct_sparse} or restrict $z_j$ to be on the training data \cite{Keerthi06buildingsupport}.
These methods essentially exploit low rank approximations of the gram matrix $\mathbf{K}$.

Low rank approximations of PD $\mathbf{K} \succ 0$, result in speedups in training and testing complexities of kernelized SVM.
Methods that learn basis coordinates outside the training data \eg \cite{direct_sparse}, usually involve intermediate optimization overheads, and thus are prohibitive in large scale scenarios.
On the contrary, the Nystr\"{o}m method gives a low rank PSD approximation to $\mathbf{K}$ with a very low cost.

The Nystr\"{o}m method \cite{Williams00theeffect} approximates $\mathbf{K}$ using a randomly selected subset of the data: 
\begin{equation}
	\mathbf{K} \approx \mathbf{K}_{\mathbf{{n}}\mathbf{{m}}} \mathbf{K}_{\mathbf{{m}}\mathbf{{m}}} ^{-1} \mathbf{K}_{\mathbf{{m}}\mathbf{{n}}}
\label{paperE:eqn:nystrom_low_rank}
\end{equation}
 where 
 $\mathbf{K}_{\mathbf{{a}}\mathbf{{b}}}$  refers to a sub matrix of $\mathbf{K} = \mathbf{K}_{\mathbf{{n}}\mathbf{{n}}}$ indexed by  $\mathbf{{a}}=(a_1,\ldots,a_n)^\trans, \, a_i \in \{0,1\}$, and similarly by $\mathbf{{b}}$.
The approximation (\ref{paperE:eqn:nystrom_low_rank}) is derived by defining eigenfunctions of $k(.,.)$ as expansions of numerical eigenvectors of $\mathbf{K}$.
A consequence is that the data can be embedded in an Euclidean space:
$\mathbf{K} \approx \mathbf{\Psi}_{\mathbf{{m}}\mathbf{{n}}}^\trans\mathbf{\Psi}_{\mathbf{{m}}\mathbf{{n}}}$, 
where $\mathbf{\Psi}_{\mathbf{{m}}\mathbf{{n}}}$, the Nystr\"{o}m feature space, is 
\begin{equation}
	\mathbf{\Psi}_{\mathbf{{m}}\mathbf{{n}}}=\mathbf{K}_{\mathbf{{m}}\mathbf{{m}}}^{-\frac{1}{2}}\mathbf{K}_{\mathbf{{m}}\mathbf{{n}}}
\label{paperE:eqn:nystrom_feature_map}
\end{equation}
Methods exist which either explicitly or implicitly exploit this \eg \cite{Lee01rsvm:reduced} to reduce both the training and test costs, by restricting the support vectors to be a subset of 
the bases defined by $\mathbf{m}$.

In case of indefinite similarity measures, $\mathbf{K}_{\mathbf{{m}}\mathbf{{m}}}^{-\frac{1}{2}}$ in (\ref{paperE:eqn:nystrom_feature_map}) will not be real.
In the rest of the paper, we refer to an indefinite version of a similarity matrix $\mathbf{K}$ with $\tilde{\mathbf{K}}$, and refer to the normalization by $\mathbf{K}_{\mathbf{{m}}\mathbf{{m}}}^{-\frac{1}{2}}$ with Nystr\"{o}m normalization.
In order to get a PSD approximation of an indefinite $\tilde{\mathbf{K}}$, the indefinite $\tilde{\mathbf{K}}_{\mathbf{{m}}\mathbf{{m}}}$ (\ref{paperE:eqn:nystrom_low_rank}) needs to be made PSD. 
Spectrum clipping, spectrum flip, spectrum shift, and spectrum square are possible solutions based on eigen decomposition of $\tilde{\mathbf{K}}_{\mathbf{{m}}\mathbf{{m}}}$.
The latter can be achieved without the eigen decomposition step: $\tilde{\mathbf{K}}_{\mathbf{{m}}\mathbf{{m}}}^\trans\tilde{\mathbf{K}}_{\mathbf{{m}}\mathbf{{m}}} \succeq 0$.

If the goal is to find the PSD matrix closest to the original indefinite $\tilde{\mathbf{K}}$ with respect to the reduced basis set $\mathbf{m}$, spectrum clip gives the closed form solution.
Therefore, when there are a few negative eigenvalues, the spectrum clip technique gives good low rank approximations to $\tilde{\mathbf{K}}_{\mathbf{{m}}\mathbf{{m}}}$ which can be used by (\ref{paperE:eqn:nystrom_low_rank}) to get a low rank PSD approximation to $\tilde{\mathbf{K}}$.
However, when there are a considerable number of negative eigenvalues, as it is the case with most of the similarity measures we consider later on in section \ref{paperE:sec:model:similarity_measures},
there is no guarantee for the resulting PSD matrix to be optimal for classification.
This is true specially when eigenvectors associated with negative eigenvalues contain discriminative information.
We experimentally verify in section \ref{paperE:sec:experiments:invariant} that the negative eigenvalues do contain discriminative information.

We seek normalizations that do not assume a PSD $\mathbf{K}_{\mathbf{m}\mathbf{m}}$, and do not require eigen-decompositions.
For example, one can replace $\mathbf{K}_{\mathbf{{m}}\mathbf{{m}}}$ in (\ref{paperE:eqn:nystrom_feature_map}) with the covariance of columns of $\mathbf{K}_{\mathbf{{m}}\mathbf{{n}}}$. 
We experimentally found out that a simple embedding, presented in the next section in (\ref{paperE:eqn:besvm_feature_map_norm}),
is competitive with the Nystr\"{o}m embedding (\ref{paperE:eqn:nystrom_feature_map}) for PSD similarity measures,
while outperforming it in case of indefinite ones that we studied.

\subsection{Basis Expanding SVM}
\label{paperE:sec:model:be_svm}
Basis Expanding SVM (BE-SVM) is a linear SVM classifier equipped with a normalization of the following explicit feature map 
\begin{equation}
	\tilde\varphi(\mathbf{x}) = \left[ s(\mathbf{b}_1,\mathbf{x}), \ldots, s(\mathbf{b}_B,\mathbf{x})\right]^\trans
\label{paperE:eqn:besvm_feature_map}
\end{equation}
where $\mathcal{B} = \{\mathbf{b}_1,\ldots, \mathbf{b}_B\}$ is an ordered basis set\footnote{For the moment assume $\mathcal{B}$ is given. We experiment with different basis selection strategies in section \ref{paperE:sec:experiments:basis_sel}).} which is a subset of the training data, and $s(.,.)$ is a pairwise similarity measure. 
The BE-SVM feature space defined by 
\begin{equation}
	\varphi(\mathbf{x}) = \frac{1}{\mathbb{E}_\mathcal{X}[\|\tilde\varphi - \mathbb{E}_\mathcal{X}[\tilde\varphi] \|]} \left(\tilde\varphi(\mathbf{x}) - \mathbb{E}_\mathcal{X}[\tilde\varphi] \right)
\label{paperE:eqn:besvm_feature_map_norm}
\end{equation}
is similar to the Nystr\"{o}m feature space (\ref{paperE:eqn:nystrom_feature_map}) with a different normalization scheme, as pointed out in section \ref{paperE:sec:model:speedup}.
The centralization of $\tilde\varphi(.)$ better conditions $\varphi(.)$ for a linear SVM solver, and normalization by the average $\ell_2$ norm is most useful for combining multiple similarity measures.

The BE-SVM classifier is 
\begin{equation}f_\mathcal{B}(\mathbf{x}) = \mathbf{w}^\trans \varphi(\mathbf{x}) \geq 0\end{equation}
where $\mathbf{w}$ is solved by minimizing the primal BE-SVM objective
\begin{equation} 
\frac{1}{2}\|\mathbf{w}\|_2^2 + C\sum_i \ell_H(y_i,f_\mathcal{B}(\mathbf{x}_i))^2 
\label{paperE:besvm_primal}
\end{equation}
An $\ell_1$ regularizer results in sparser solutions, but with the cost of more expensive optimization than an $\ell_2$ regularization.
Therefore, for large scale scenarios, an $\ell_2$ regularization, combined with a reduced basis set $\mathcal{B}$, is preferred to an $\ell_1$ regularizer combined with a larger basis set.

Using multiple similarity measures is straightforward in BE-SVM.
The concatenated feature map $\varphi_M(\mathbf{x}) = \left[ \varphi^{(1)}(\mathbf{x})^\trans, \ldots, \varphi^{(M)}(\mathbf{x})^\trans\right]^\trans$ encodes the values of the $M$ similarity measures evaluated on the corresponding bases $\mathcal{B}^{(1)},\ldots,\mathcal{B}^{(M)}$. 
In this work, we restrict the study to the case that the bases are shared among the $M$ similarity measures: \ie $\mathcal{B}^{(1)}=\ldots=\mathcal{B}^{(M)}$.

\subsection{Indefinite Similarity Measures for Visual Recognition}
\label{paperE:sec:model:similarity_measures}
The lack of expressibility of the PSD kernels have been argued before \eg in \cite{sim_class,learning_kernels_from_indef,good_rec_nonmetric}.
For example, similarity measures which are not based on vectorial representations of data are most likely to be indefinite.
Particularly in computer vision, considering latent information results in lack of a fixed vectorial representation of instances, and therefore similarity measures based on latent information are most likely to be indefinite\footnote{Note that \cite{YangWVM12} and similar approaches use a PD kernel on fixed vectorial representation of the data, \emph{given the latent information}. The latent informations in turn are updated using an alterantive minimization approach. This makes the optimization non-convex, and differs from similarity measures which directly model latent informations.}.

A few applications of indefinite similarity measures in computer vision are pointed out below.
\cite{invariant_kernels} proposes (indefinite) jitter kernels for building desired invariances in classification problems.
\cite{AghazadehASC12} uses indefinite pairwise similarity measures with latent positions of objects for clustering.
\cite{deformable_models} considers deformation models for image matching.
\cite{graph_matching} defines an indefinite similarity measure based on explicit correspondences between pairs of images for image classification.

In this work, we consider similarity measures with latent deformations:
\begin{equation}
s(x_i,x_j) = \underset{z_i \in \mathcal{Z}(x_i),\,z_j \in \mathcal{Z}(x_j)}{\max}  K_I(\mathbf{\phi}(x_i,z_i), \mathbf{\phi}(x_j,z_j)) +\mathcal{R}(z_i) + \mathcal{R}(z_j)  
\label{paperE:eqn:indef_sim}
\end{equation}
where $K_I(.,.)$ is a similarity measure (potentially a PD kernel), $\mathbf{\phi}(x,z)$ is a representation of $x$ given the latent variable $z$, $\mathcal{R}(z)$ is a regularization term on the latent variable $z$, and $\mathcal{Z}(x)$ is the set of possible latent variables associated with $x$.
Specifically, when $\mathcal{R}(.) = 0$ and $\mathcal{Z}(x)$ involves latent positions, the similarity measure becomes similar to that of \cite{AghazadehASC12}.
When $\mathcal{R}(.) = 0$ and $\mathcal{Z}(x)$ involves latent positions and local deformations, it becomes similar to the zero order model of \cite{deformable_models}.
Finally, an MRF prior in combination with latent positions and local deformations gives a similarity measure, similar to that of \cite{graph_matching}.

The proposed similarity measure (\ref{paperE:eqn:indef_sim}) picks the latent variables which have the maximal (regularized) similarity values $K_I(.,.)$s. 
This is in contrast to \cite{invariant_kernels} where the latent variables were suggested to be those which minimize a metric distance based on the kernel $K_I(.,.)$.
The advantage of a metric based latent variable selection is not so clear, while some works argue against unnecessary restrictions to metrics \cite{good_rec_nonmetric}.
Also, if $K_I(.,.)$ is not PSD, deriving a metric from it is at best expensive. 
Therefore, the latent variables in (\ref{paperE:eqn:indef_sim}) are selected according to the similarity values instead of metric distances.

\subsection{Multi Class Classification}
\label{paperE:sec:model:multi_class}
SVMs are mostly known as binary classifiers.
Two popular extensions to the multi-class problems are one-v-res (1vR) and one-v-one (1v1). 
The two simple extentions have been argued to perform as well as more sophisticated formulations \cite{defense_onevall}.
In particular, \cite{defense_onevall} concludes that in case of kernelized SVMs, in terms of accuracy they are both competitive, while in terms of training and testing complexities 1v1 is superior.
Therefore, we only consider 1v1 approach for kernelized SVM.
In case of linear SVMs however, 1v1 results in unnecessary overhead and 1vR is the algorithm of choice.
A 1vR BE-SVM can be expected to be both faster and to generalize better than a 1v1 BE-SVM where bases from all classes are used in each of the binary classifiers.
In case of 1v1 BE-SVM where only bases from the two classes under consideration are used in each binary classifier, there will be a clear advantage in terms of training complexity.
However, due to the reduction in the size of the basis set, the algorithm generalizes less in comparison to a 1vR approach.
Therefore, we only consider 1vR formulation for BE-SVM.
Table \ref{paperE:tab:complexity} summarizes the memory and computational complexity analysis for 1v1 kernelized SVM and 1vR BE-SVM.
Shown are the upper bounds complexities where we have considered $n$ to be the upper bound on $n_{sv}$.

\begin{table}
\centering 
\begin{tabular}{|c||c|c|c|c|}
\hline 
& \multicolumn{2}{|c|}{Training} & \multicolumn{2}{|c|}{Testing (per sample)} \\ \hline
& Memory  & Computation & Memory  & Computation \\ \hline
K SVM 	&  $nM\bar{d}_\mathbf{\phi} + \frac{n^2}{C}$& $n^2M\bar{d}_K + \frac{n^3}{C}$ & $nM\bar{d}_\mathbf{\phi}$ & $nCM\bar{d}_K$\\ \hline
BE-SVM 	& $nM\bar{d}_\mathbf{\phi}+ nM|\mathcal{B}|$ & $nC|\mathcal{B}|M\bar{d}_K$ & $|\mathcal{B}|M\bar{d}_\mathbf{\phi}$ & $|\mathcal{B}|CM\bar{d}_K$ \\ \hline
\end{tabular}
\caption{Complexity Analysis for kernelized SVM and BE-SVM. The number of samples for each of the $C$ classes was assumed to be equal to $\frac{n}{C}$. $M$ is the number of kernels/similarity measures, $M\bar{d}_\mathbf{\phi}$ is the dimensionality of representations required for evaluating $M$ kernels/similarity measures, and $M\bar{d}_K$ is the cost of evaluating all $M$ kernels/similarity measures.}
\label{paperE:tab:complexity}
\end{table}

\subsection{Margin Analysis of Basis Expanding SVM}
\label{paperE:SM:sec:margin}
Both kernelized SVM and BE-SVM are max margin classifiers in their feature spaces.
The feature space of kernelized SVM $\psi_\mathcal{H}(.)$ is implicitly defined via the kernel function $k(.,.)$ while the feature space of the BE-SVM is explicitly defined via empirical kernel maps.
In order to derive the margin as a function of the data, we first need to derive the dual BE-SVM objective, where we assume a non-squared Hinge loss and unnormalized feature mappings $\tilde\varphi(.)$.
Borrowing from the representer theorem and considering the KKT conditions of the primal, 
one can derive $\mathbf{w} = \sum_i y_i \beta_i \mathbf{\tilde\varphi}(\mathbf{x}_i)$, and consequently derive the BE-SVM dual objective which is similar to the dual SVM objective but with $\mathbf{K}_{ij} = \mathbf{\tilde\varphi}(\mathbf{x}_i)^\trans \mathbf{\tilde\varphi}(\mathbf{x}_j)$.
Let $\mathbf{S}_{BX}$ refer to the similarity of the data to the bases.
We can see that the margin of the BE-SVM, given the optimal dual variables $0\leq\mathbf{\beta}_i\leq C$, is $\left(\mathbf{\beta}^\trans\mathbf{Y}\mathbf{S}_{BX}^\trans\mathbf{S}_{BX}\mathbf{Y}\mathbf{\beta}\right)^{-1}$, as opposed to $\left(\mathbf{\alpha}^\trans\mathbf{Y}\mathbf{K}\mathbf{Y}\mathbf{\alpha}\right)^{-1}$ for the kernelized SVM, given the optimal dual variables $0\leq\mathbf{\alpha}_i\leq C$. 
Furthermore, $\mathbf{S}_{BX}^\trans\mathbf{S}_{BX}$ is PSD, and that is BE-SVM's workaround for using indefinite similarity measures.

We analyze the margin of BE-SVM in case of unnormalized features ($\tilde\varphi(.)$ instead of $\varphi(.)$) and a non-squared Hinge loss.
Given the corresponding dual variables, the margin of the BE-SVM was mentioned to be 
\begin{equation}
M_{BE}(\beta) = \left(\mathbf{\beta}^\trans\mathbf{Y}\mathbf{S}_{BX}^\trans\mathbf{S}_{BX}\mathbf{Y}\mathbf{\beta}\right)^{-1}
\end{equation}
as opposed to that of the kernelized SVM
\begin{equation}
M_{K}(\alpha) = \left(\mathbf{\alpha}^\trans\mathbf{Y}\mathbf{K}\mathbf{Y}\mathbf{\alpha}\right)^{-1}
\end{equation}
For comparison, the margin of the Nystr\"{o}mized method is
\begin{equation}
M_{N}(\alpha) = \left(\mathbf{\alpha}^\trans\mathbf{Y}\mathbf{K}_{XB}\mathbf{K}_{BB}^{-1}\mathbf{K}_{BX}\mathbf{Y}\mathbf{\alpha}\right)^{-1}
\end{equation}

\noindent\textbf{BE-SVM vs Kernelized SVM}:
When $s(.,.) = k(.,.)$ and all training exemplars are used as bases, the margin of the BE-SVM will be $\left(\mathbf{\beta}^\trans\mathbf{Y}\mathbf{K}^2\mathbf{Y}\mathbf{\beta}\right)^{-1}$.
Comparing to the margin of SVM, for the same parameter $C$ and the same kernel, it can be said that the solution (and thus the margin) of BE-SVM is even more derived by large eigenpairs, and even less by small ones.
It is straightforward to verify $\mathbf{K}^2=\sum_i \lambda_i^2 \mathbf{v}_i \mathbf{v}_i^\trans$.
Therefore, the contribution of large eigenpairs, that are $\{(\lambda_i,\mathbf{v}_i)|\lambda_i>1\}$, to $\mathbf{K}^2$ is amplified.
Similarly, the contribution of small eigenpairs, that are those with $\lambda_i<1$, to $\mathbf{K}^2$ is dampened.

\noindent\textbf{BE-SVM vs Nystr\"{o}mized method}:
When $s(.,.) = k(.,.)$ and a subset of training exemplars are used as bases (reduced settings), the resulting margin of BE-SVM is $\left(\mathbf{\beta}^\trans\mathbf{Y}\mathbf{K}_{XB}\mathbf{K}_{BX}\mathbf{Y}\mathbf{\beta}\right)^{-1}$.
Comparing to the margin of the Nystr\"{o}mized method, we can say that the most of the difference between the Nystr\"{o}mized method and BE-SVM, is the normalization by $\mathbf{K}_{BB}^{-1}$.

For covariance kernels, that the Nystr\"{o}mized method is most suitable for, $\mathbf{K}_{BB}$ is the covariance of the basis set in the feature space. 
Therefore, it can be said that the normalization by $\mathbf{K}_{BB}^{-1}$ essentially de-correlates the bases in the feature space.
Although this is an appealing property, there is no associating RKHS with indefinite similarity measures and the de-correlation in such cases is non-trivial.
In case of covariance kernels, it can be said that BE-SVM assumes un-correlated bases, while bases are always correlated in the feature space.
As larger sets of bases usually result in more (non-diagonal) covariances, the un-correlated assumption is more violated with large set of bases.
The consequence is that in such cases, that are covariance kernels with large set of bases, BE-SVM can be expected to perform worse than the Nystr\"{o}mized method.
However, for sufficiently small set of bases, or in case of indefinite similarity measures, there is no reason for superiority of the Nystr\"{o}mized method.
In such cases and in practice, BE-SVM is competitive or better than the Nystr\"{o}mized method.

\subsubsection{Demonstration on 2D Toy data}
\label{paperE:SM:sec:margin:2D}
\begin{figure}
\centering
\begin{subfigure}[b]{0.32\textwidth}
	\includegraphics[width=\textwidth]{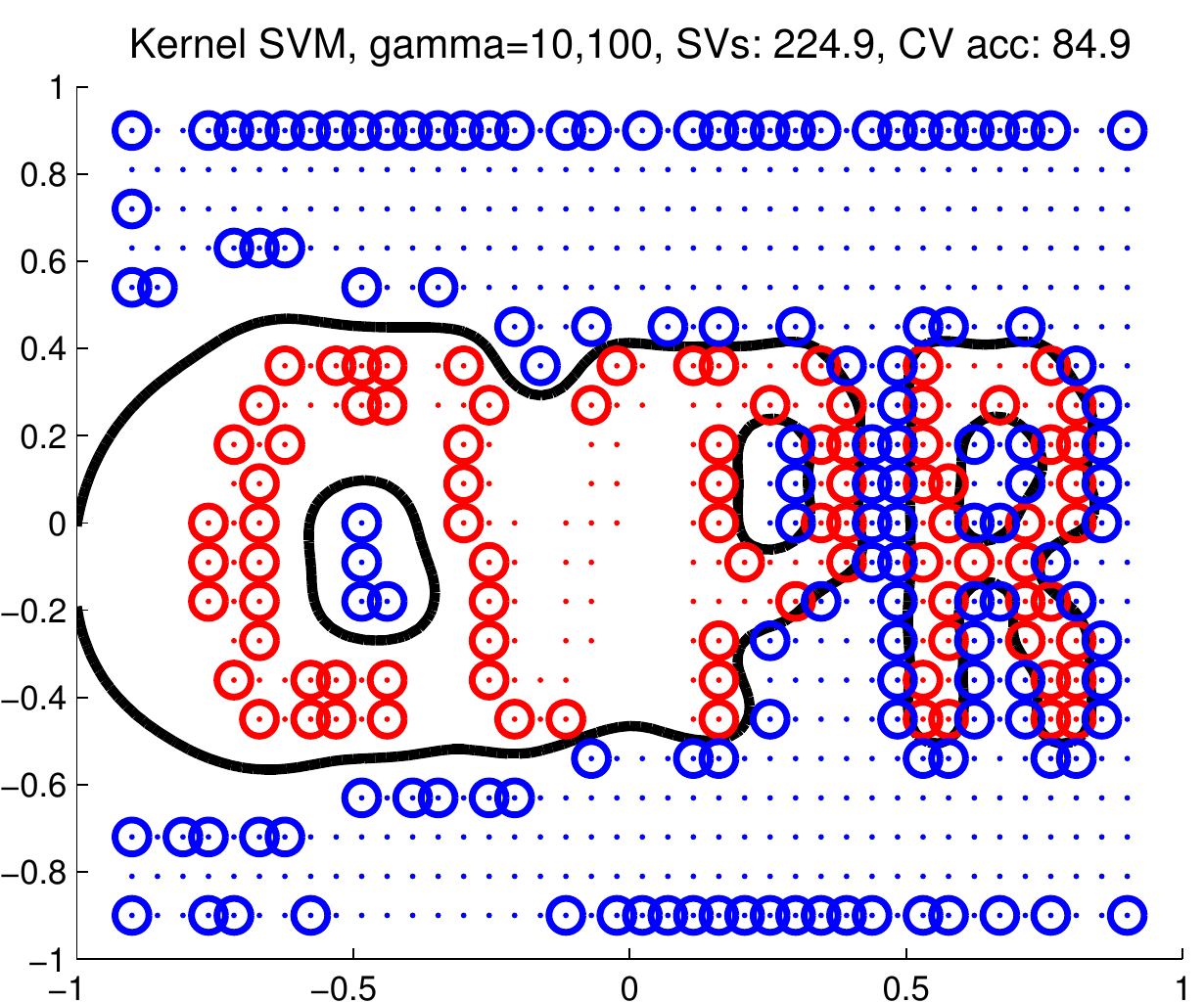} 
	\caption{Kernelized SVM}
	\label{paperE:SM:fig:besvm_ksvm_2D:ksvm}
\end{subfigure}
\begin{subfigure}[b]{0.32\textwidth}
	\includegraphics[width=\textwidth]{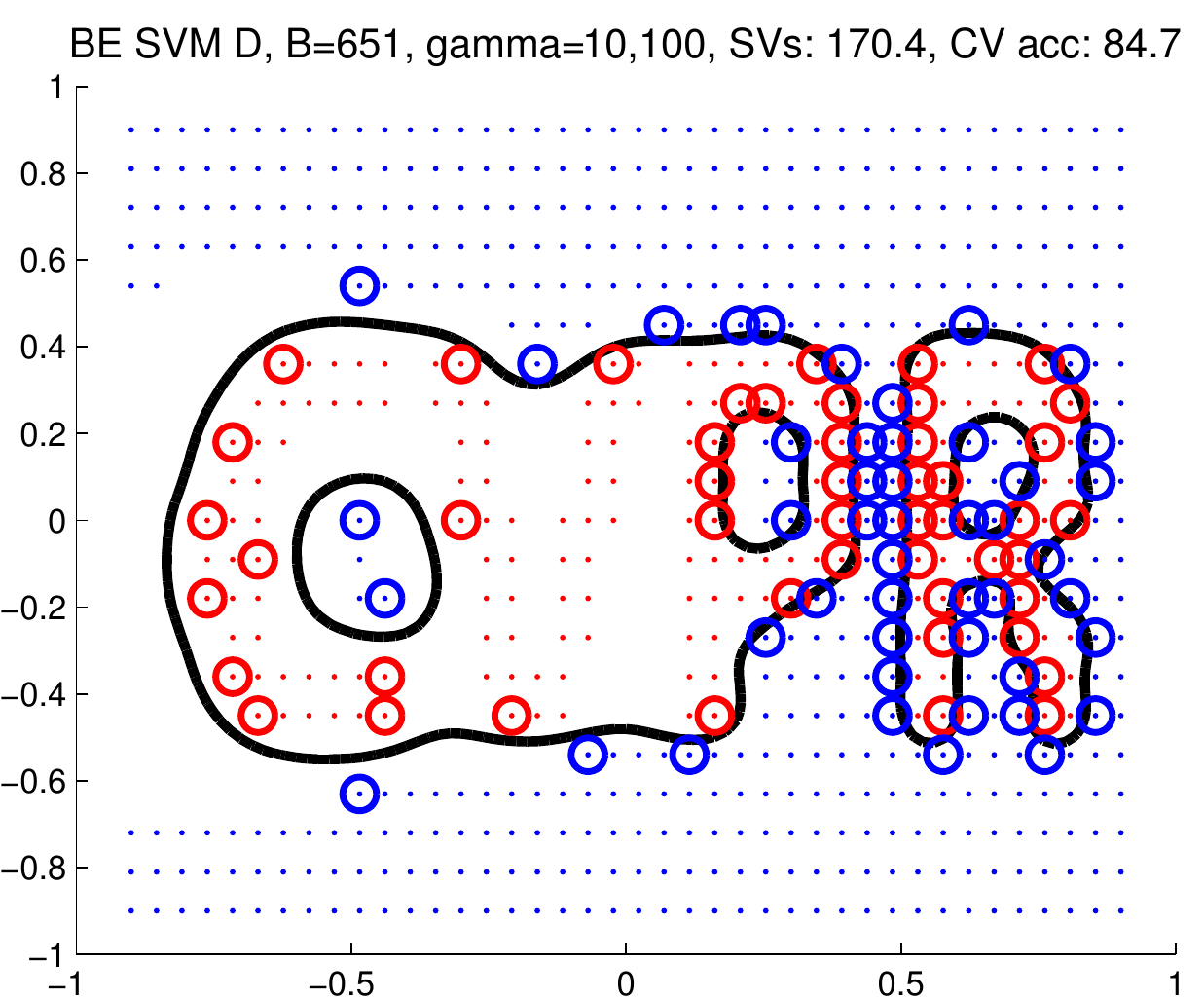} 
	\caption{BE-SVM dual objective}
	\label{paperE:SM:fig:besvm_ksvm_2D:dual}
\end{subfigure}
\begin{subfigure}[b]{0.32\textwidth}
	\includegraphics[width=\textwidth]{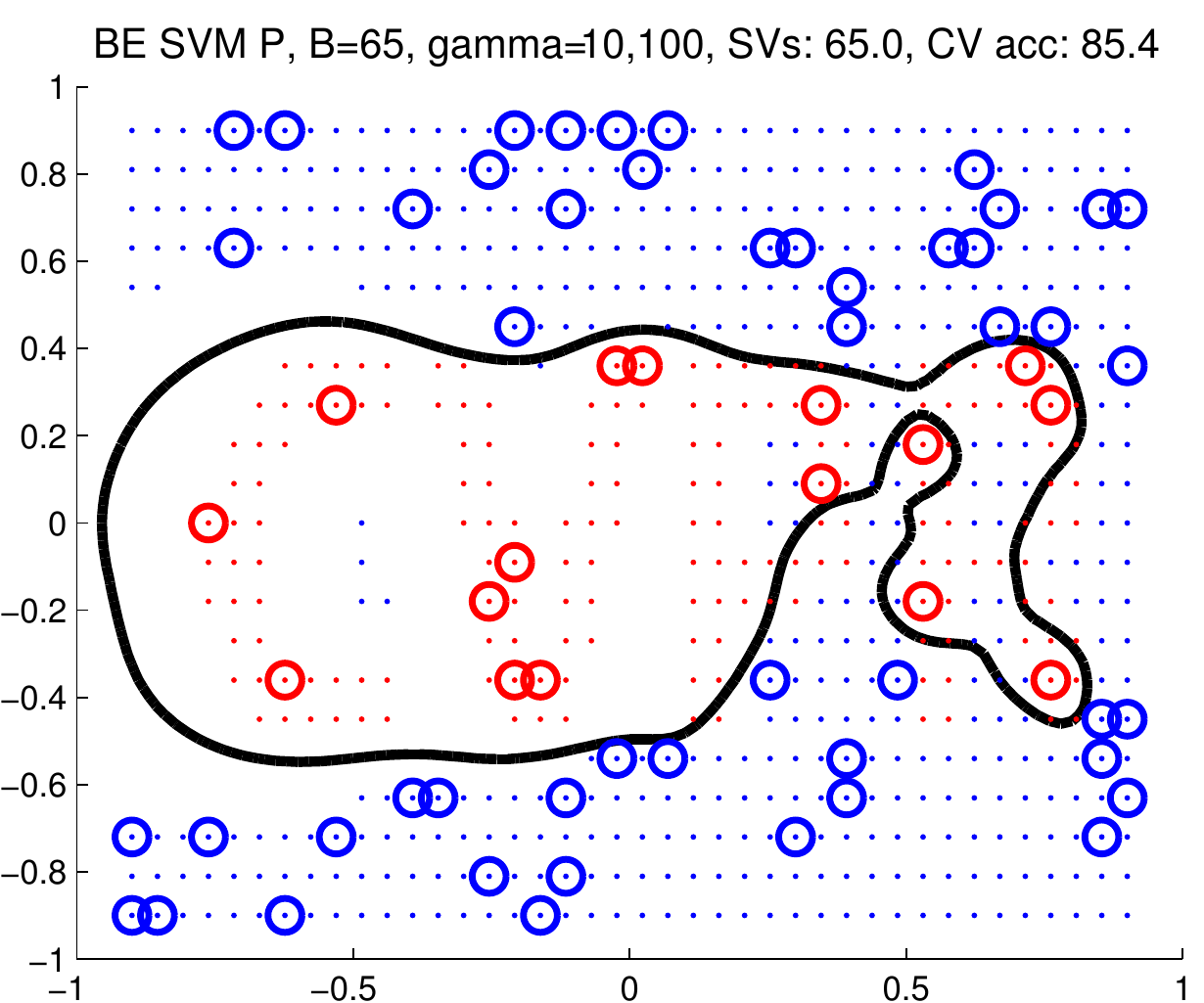} 
	\caption{BE-SVM primal objective}
	\label{paperE:SM:fig:besvm_ksvm_2D:primal_reduced}
\end{subfigure}
\caption{Demonstration of kernelized SVM and BE-SVM using two Gaussian RBF kernels with $\gamma_1=10,\gamma_2=10^2$ and $C=10$. 
\ref{paperE:SM:fig:besvm_ksvm_2D:ksvm} is based on equally weighted kernels. 
\ref{paperE:SM:fig:besvm_ksvm_2D:dual} is without normalization. 
\ref{paperE:SM:fig:besvm_ksvm_2D:primal_reduced} is with normalization on 10\% of the data randomly selected as bases. 
10 fold cross validation accuracy and the number of support vectors are averaged over $i=1:20$ scenarios based on the same problem but with different spatial noises. 
The noise model for $i^\text{th}$ scenario is a zero mean Gaussian with $\sigma_i=10^{-2}i$. 
The visualization is on the noiseless data for clarity. 
Best viewed electronically.}
\label{paperE:SM:fig:besvm_ksvm_2D}
\end{figure}

Figure \ref{paperE:SM:fig:besvm_ksvm_2D} visualizes the use of multiple Gaussian RBF kernels in BE-SVM and kernelized SVM. 
We point out the following observations.

1) the dual objective of BE-SVM (exact) tends to result in sparser solutions as measured by non-zero support vector coefficient (compare \ref{paperE:SM:fig:besvm_ksvm_2D:ksvm} with \ref{paperE:SM:fig:besvm_ksvm_2D:dual}). 
We believe the main reason for this to be the modification of the eigenvalues as described in section \ref{paperE:SM:sec:margin}.
Note however that in order to classify a new sample, its similarity to all training data needs to be evaluated, irrespective of the sparsity of the BE-SVM solution (see equation (\ref{paperE:eqn:besvm_mult_kernel})). 
In this sense, the BE-SVM dual objective results in completely dense solutions, similar to the primal BE-SVM objective without any basis reduction.
However, the solution can be made sparse by construction, by reducing the basis set, similar to the case with the primal BE-SVM objective.
We do not demonstrate this here, mainly because our main focus is on the (approximate) primal objective.

2) due to the definition of the (linear) kernel in BE-SVM (see equation (\ref{paperE:eqn:besvm_mult_kernel})), the solution of the BE-SVM has an inherent bias with respect to the (marginal) distribution of class labels.
In other words, the contribution of each class to the norm of $\tilde\varphi(.)$, and consequently to the value of $\tilde{K}(.,.)$, directly depends on the number of bases from each class.
Consequently, the decision boundary of BE-SVM is shifted towards the class with less bases: compare the decision boundaries on the left sides of \ref{paperE:SM:fig:besvm_ksvm_2D:ksvm} and \ref{paperE:SM:fig:besvm_ksvm_2D:dual}. 
In experiments on CIFAR-10 dataset, as the number of exemplars from different classes are roughly equal, this did not play a crucial role.

\subsection{Related Work}
\label{paperE:sec:model:related_work}
There exists a body of work regarding the use of proximity data, similarity, or dissimilarity measures in classification problems.
\cite{similarity_svm} uses similarity to a fixed set of samples as features for a kernel SVM classifier.
\cite{pairwise_proximity} uses proximities to all the data as features for a linear SVM classifier. 
\cite{classification_proximity_lp} uses proximities to all the data as features and proposes a linear program machine based on this representation.
In contrast, we use a normalization of the similarity of points to a subset of the data as features for a (fast, approximate) linear SVM classifier.

\section{Experiments}
\label{paperE:sec:experiments}

\subsection{Dataset and Experimental Setup}
\label{paperE:sec:experiments:dataset}
We present our experimental results on CIFAR-10 dataset \cite{Krizhevsky09learningmultiple}.
The dataset is comprised of 60,000 tiny $32\times32$ RGB images, 6,000 images for each of the 10 classes involved, divided into 6 folds with inequal distribution of class labels per fold.
The first 5 folds are used for training and the 6th fold is used for testing.
We use a modified version of the HOG feature \cite{DT05}, described in \cite{FelzenszwalbGMR10}.
For most of our experiments, we use HOG cell sizes of 8 and 4, which result in $31\times\frac{32}{8}^2=496$ and $31\times\frac{32}{4}^2=1984$ dimensional representation of each of the images.

Due to the normalization of each of the HOG cells, namely normalizing by gradient/contrast information of the neighboring cells, the HOG cells on border of images are not normalized properly. 
We believe this to have a negative effect on the results, but as the aim of this paper is not to get the best results possible out of the model, we rely on the consistency of the normalization for all images to address this problem. 
A possible fix is to up-sample images and ignore the HOG-cells at the boundaries, but we do not provide the results for such fixes.

For all the experiments, we center the HOG feature vector and scale feature vectors inversely by the average $\ell_2$ norm of the centered feature vectors, similar to the normalization of BE-SVM (\ref{paperE:eqn:besvm_feature_map_norm}).
This results in easier selection of parameters $C$ and $\gamma$ for SVM formulations. 
Unless stated otherwise, we fix $C=2$ and $\gamma=1$ for kernelized SVM with Gaussian RBF kernels, and $C=1$ for the rest.
We use LibLinear \cite{REF08a} to optimize the primal linear SVM objectives with squared Hinge loss, similar to (\ref{paperE:besvm_primal}).
For kernelized SVM, we use LibSVM \cite{CC01a}.
We report multi-class classification results (0-1 loss) on the test set, where we used a 1 v 1 formulation for kernelized SVM, and 1 v all formulation for other methods.

\subsection{Baseline: SVM with Positive Definite Kernels}
Figure \ref{paperE:fig:multiple_sims_params_support:params} shows  the performance of linear SVM (H4L and H8L) and kernelized SVM with Gaussian RBF kernel (K4R and K8R) as a function of number of parameters in the models.
The number of parameters for linear SVM is the input dimensionality, and for kernelized SVM it is the sum of $n_{sv} (d_\phi+1)$ where $d_\phi$ is the dimensionality of the feature vector the corresponding kernel operates on.
The 5 numbers for each model are the results of the model trained on $1,\ldots,5$ folds of the training data (each fold contains 10,000 samples).
Figure \ref{paperE:fig:multiple_sims_params_support:support} shows the performance kernelized SVM as a function of support vectors when trained on $1,\ldots,5$ folds.
Except the linear SVM with a HOG cell size of 8 pixels (496 dimensions) which saturates its performance at 4 folds, all models consistently benefit from more training data.

\subsection{BE-SVM with Invariant Similarity Measures}
\label{paperE:sec:experiments:invariant}
The general form of the invariant similarity measures we consider was given in (\ref{paperE:eqn:indef_sim}).
In particular, we consider rigid and deformable similarity measures where the smallest unit of deformation/translation is a HOG cell.

The rigid similarity measure models invariance to translations and is given by
\begin{equation}
K_R(x,y) = \max_{\mathbf{z}_R \in \mathcal{Z}_R} \sum_{\mathbf{c} \in \mathcal{C}} \phi_C(x,\mathbf{c})^T \phi_C(y,\mathbf{c} + \mathbf{z}_R)
\label{paperE:eqn:sim:rigid}
\end{equation}
where $\mathcal{Z}_R =\{(z_x,z_y)|z_x,z_y \in \{-h_R,\ldots,h_R\}\}$ allows a maximum of $h_R$ HOG cells displacements in $x,y$ directions, $\mathcal{C} = \{(x,y)|x,y \in \{h_1,\ldots,h_H\}$ is the set of indices of $h_H$ HOG cells in each direction, and $\phi_C(x,\mathbf{c})$ is the 31 dimensional HOG cell of $x$ located at position $\mathbf{c}$.
$\phi_C(x,\mathbf{c})$ is zero for cells outside $x$ (zero-padding).
$K_R(x,y)$ is the maximal cross correlation between $\phi(x)$ and $\phi(y)$. 

The deformable similarity measure allows local deformations (displacements) of each of the HOG cells, in addition to invariance to rigid deformations
\begin{equation}
K_L(x,y) = \max_{z_R \in \mathcal{Z}_R} \sum_{\mathbf{c} \in \mathcal{C}} \max_{z_L \in \mathcal{Z}_L} \phi_C(x,\mathbf{c})^T \phi_C(y,\mathbf{c} + \mathbf{z}_R + \mathbf{z}_L)
\label{paperE:eqn:sim:nonrigid}
\end{equation}
where $\mathcal{Z}_L=\{(z_x,z_y)|z_x,z_y \in \{-h_L,\ldots,h_L\}\}$ allows a maximum of $h_L$ HOG cell local deformation for each of the HOG cells of $y$.

We consider a maximum deformation of 8 pixels \eg 2 HOG cells for a HOG cell size of 4 pixels.
Regularizing global or local deformations is straightforward in this formulation. 
However, we did not notice significant improvements for the set of displacements we considered, which is probably related to the small size of the latent set suitable for small images in CIFAR-10.

\begin{figure}[t]
\centering
	\begin{subfigure}[b]{0.49\textwidth}
		\includegraphics[width=\textwidth]{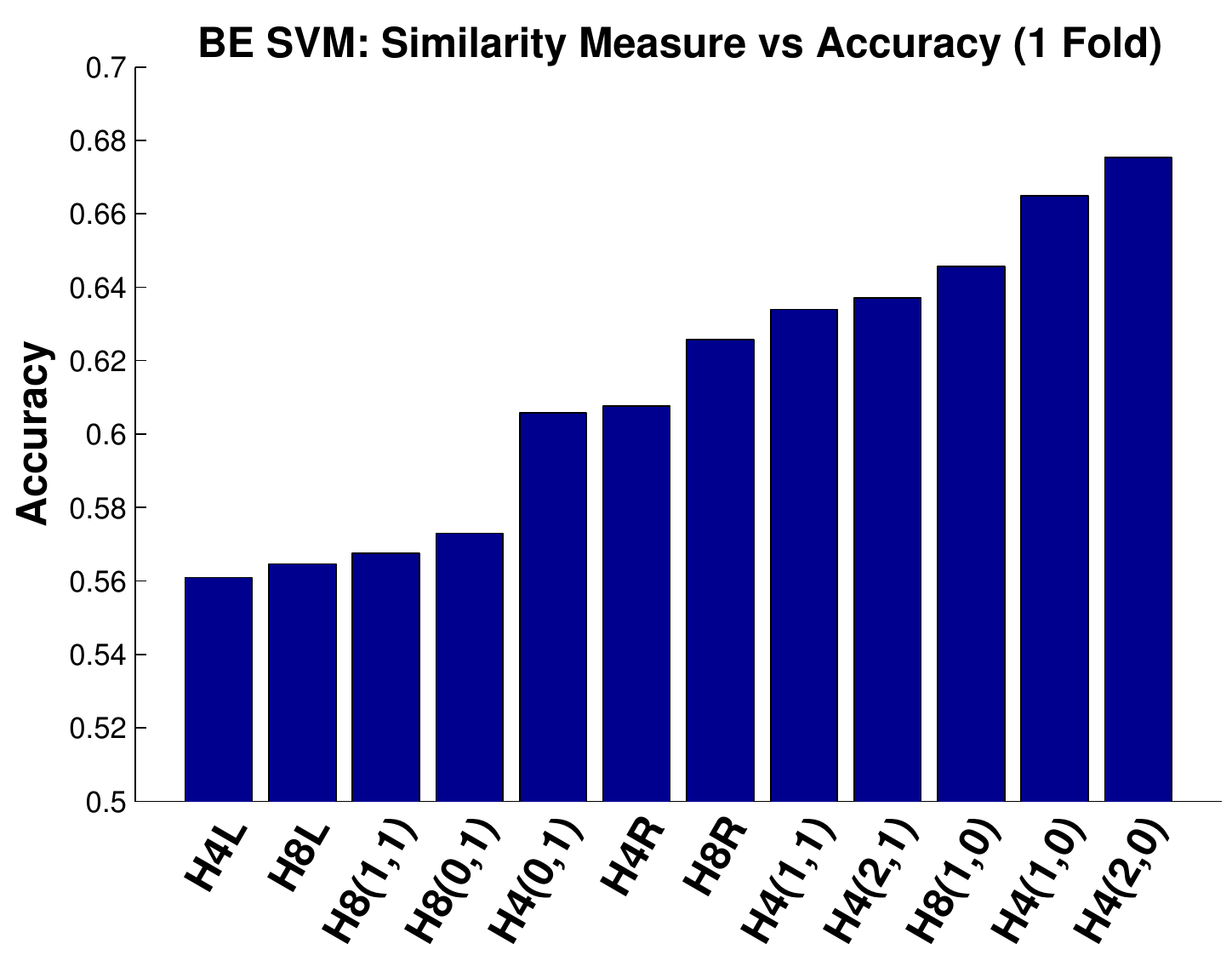}
		\caption{Full Bases}
		\label{paperE:fig:full_basis_sel:full}
	\end{subfigure}
	\begin{subfigure}[b]{0.49\textwidth}
		\includegraphics[width=\textwidth]{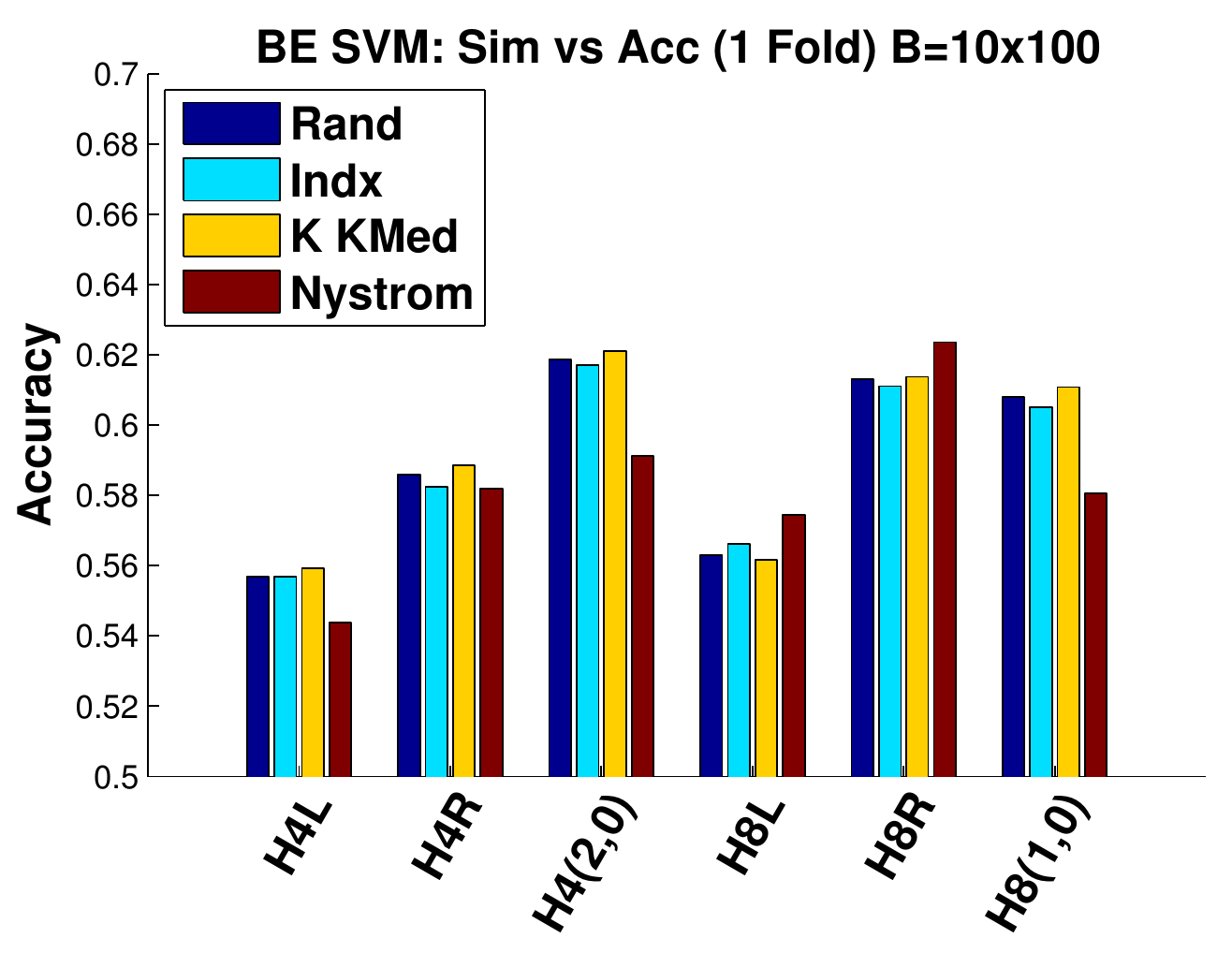}
		\caption{Basis Selection}
		\label{paperE:fig:full_basis_sel:sel}
	\end{subfigure}
\caption{Performance of BE-SVM as a function of different similarity measures when trained on the first fold. An H4 (H8) refers to a HOG cell size of 4 (8) pixels. L and R refer to  linear and Gaussian RBF kernels respectively, and $(h_R,h_L)$ refers to a similarity measure with $h_R$ rigid and $h_L$ local deformations (\ref{paperE:eqn:sim:rigid}), (\ref{paperE:eqn:sim:nonrigid}).} 
\label{paperE:fig:full_basis_sel}
\end{figure}

Figure \ref{paperE:fig:full_basis_sel:full} shows the performance of BE-SVM using different similarity measures, when trained on the first fold.
It can be seen that the invariant similarity measures improve recognition performance.
Particularly, in absence of any other information, modelling rigid deformations (latent positions) seems to be much more beneficial than modelling local deformations.
An interesting observation is that aligning the data in higher resolutions is much more crucial: all models (linear SVM, kernelized SVM, and BE-SVM) suffer performace losses when the resolution is increased from a HOG cell size of 8 pixels to 4 pixels. 
However, BE-SVM achieves significant performance gains by aligning the data in higher resolutions: compare H4L with H4(1,0) and H4(2,0), and H8L with H8(1,0).

We tried training linear and kernelized SVM models by jittering the feature vectors, in the same manner that the invariant similarity measures do (\ref{paperE:eqn:sim:rigid}), (\ref{paperE:eqn:sim:nonrigid}); that is to jitter the HOG cells with zero-padding for cells outside images.
This resulted in significant performance losses for both linear SVM and kernelized SVM, while also siginificantly increasing memory requirement and computation times.
We believe the reason for this to be the boundary effects; which are also mentioned in previous work \eg \cite{invariant_kernels}.
We also believe that jittering the input images, in combination with some boundary heuristics (see section \ref{paperE:sec:experiments:dataset}), will improve the test performance (while significantly increasing training complexities), but we do not provide experimental results for such cases.

\subsection{Basis Selection}
\label{paperE:sec:experiments:basis_sel}
Figure \ref{paperE:fig:full_basis_sel:sel} shows accuracy of BE-SVM using different similarity measures and different basis selection strategies; for a basis size of $B=10\times100$ exemplars.
In the figure, `Rand' refers to a random selection of the bases, `Indx' refers to selection of samples according to their indices, `K KMed' refers to a kernel k-medoids approach based on the similarity measure, and `Nystrom' refers to selection of bases similar to the `Indx' approach, but with the Nystr\"{o}m normalization, using a spectrum clip fix for indefinite similarity measures(see section \ref{paperE:sec:model:speedup}).
The reported results for `Rand' method is averaged over 5 trials; the variance was not significant.
It can be observed that all methods except the `Nystrom' result in similar performances.
We also tried other sophisticated sample selection criteria, but observed similar behaviour. 
We attribute this to little variation in the quality of exemplars in the CIFAR-10 dataset.
Having observed this, for the rest of sub-sampling strategies, we do not average over multiple random basis selection trials, but rather use the deterministic `Indx' approach.

The difference between normalization factors in BE-SVM and Nystr\"{o}m method (see section \ref{paperE:sec:model:speedup}) is evident in the figure.
The BE-SVM normalization tends to be consistently superior in case of indefinite similarity measures.
For PSD kernels (H4L, H8L, H4R, and H8R) , the Nystr\"{o}m normalization tends to be better in lower resolutions (H8) and worse in higher resolutions (H4).
We believe the main reason for this is to be lack of significant similarity of bases in higher resolutions in absence of any alignment.
In such cases, the low rank assumption of $\mathbf{K}$ \cite{Williams00theeffect} is violated, and normalization by a diagonally dominant $\mathbf{K}_{\mathbf{{m}}\mathbf{{m}}}$ will not capture any useful information.

In order to analyze how the performance of BE-SVM depends on the eigenvalues of the similarity measures, we provide the following eigenvalue analysis.
We compute the similarity of the bases to themselves -- corresponding to $\mathbf{K}_{\mathbf{{m}}\mathbf{{m}}}$ in (\ref{paperE:eqn:nystrom_feature_map}) -- and perform an eigen-decomposition of the resulting matrix.
Table \ref{paperE:tab:eigs} shows the ratio of negative eigenvalues: `NgRat'=$\frac{1}{B}\sum_i [\lambda_i<0]$ , and the relative energy of eigenvalues `NgEng'=$\frac{\sum_i |\lambda_i|[\lambda_i <0]}{\sum_i |\lambda_i|[\lambda_i >0]}$ as a function of various similarity measures for $B=10\times 100$ and a HOG cell size of 4.
The last two columns, namely `CorNyst' and `CorBE' reflect the correlation of the measured entities -- `NgRat' and `NgEng' -- to the observed performance of BE-SVM using the Nystr\"{o}m normalization and BE-SVM normalization.
We used Pearson's $r$ to measure the extent of linear dependence between the test performances and different normalization schemes.
It can be observed that: 1) both normalization schemes have a positive correlation with both the ratio of negative eigenvalues and their relative energy, and 2) BE-SVM normalization correlates more strongly with the observed entities.
From this, we conclude that negative eigenvectors contain discriminative information and that BE-SVM's normalization is more suitable for indefinite similarity measures.
We also experimented with spectrum flip and spectrum square methods for the Nystr\"{o}m normalization, but they generally provided slightly worse results in comparison to the spectrum clip technique.
\begin{table}
\centering
\begin{tabular}{|c|c|c|c|c|c|c||c|c|}
\hline
&{\small H4(0,0)}&{\small H4(0,1)}&{\small H4(1,0)}&{\small H4(1,1)}&{\small H4(2,0)}&{\small H4(2,1)}&{\small CorNyst}&{\small CorBE}\\ \hline
NgRat&.0&.26&.18&.25&.16&.30&.20 & .61\\ \hline
NgEng&.00&.04&.05&.05&.04&.07&.33 & .73\\ \hline
\end{tabular}
\caption{Eigenvalue analysis of various similarity measures based on HOG cell size 4.
}
\label{paperE:tab:eigs}
\end{table}

\subsection{Multiple Similarity Measures}
\label{paperE:section:multiple_sims}
Different similarity measures contain complementary information.
Fortunately, BE-SVM can make use of multiple similarity measures by construction.
To demonstrate this, using one fold of training data and $B=10\times 50$, we greedily -- in an incremental way -- augmented the similarity measures with the most contributing ones.
Using this approach, we found two (ordered) sets of similarity measures with complementary information: 1) a low-resolution set $\mathcal{M}_1=\{H8R, H8(1,0),H8(0,1)\}$, and 2) a two-resolution set $\mathcal{M}_2 = \{ H8R, H4(2,0), H4(0,1), H8(1,0)\}$.
Surprisingly, the two resolution sequence resembles those of the part based models \cite{FelzenszwalbGMR10}, and multi resolution rigid models \cite{AghazadehASC12} in that the information is processed at two levels: a coarser rigid `root' level and a finer scale deformable level.

\begin{figure}[t]
\centering
	\begin{subfigure}[b]{0.49\textwidth}
		\includegraphics[width=\textwidth]{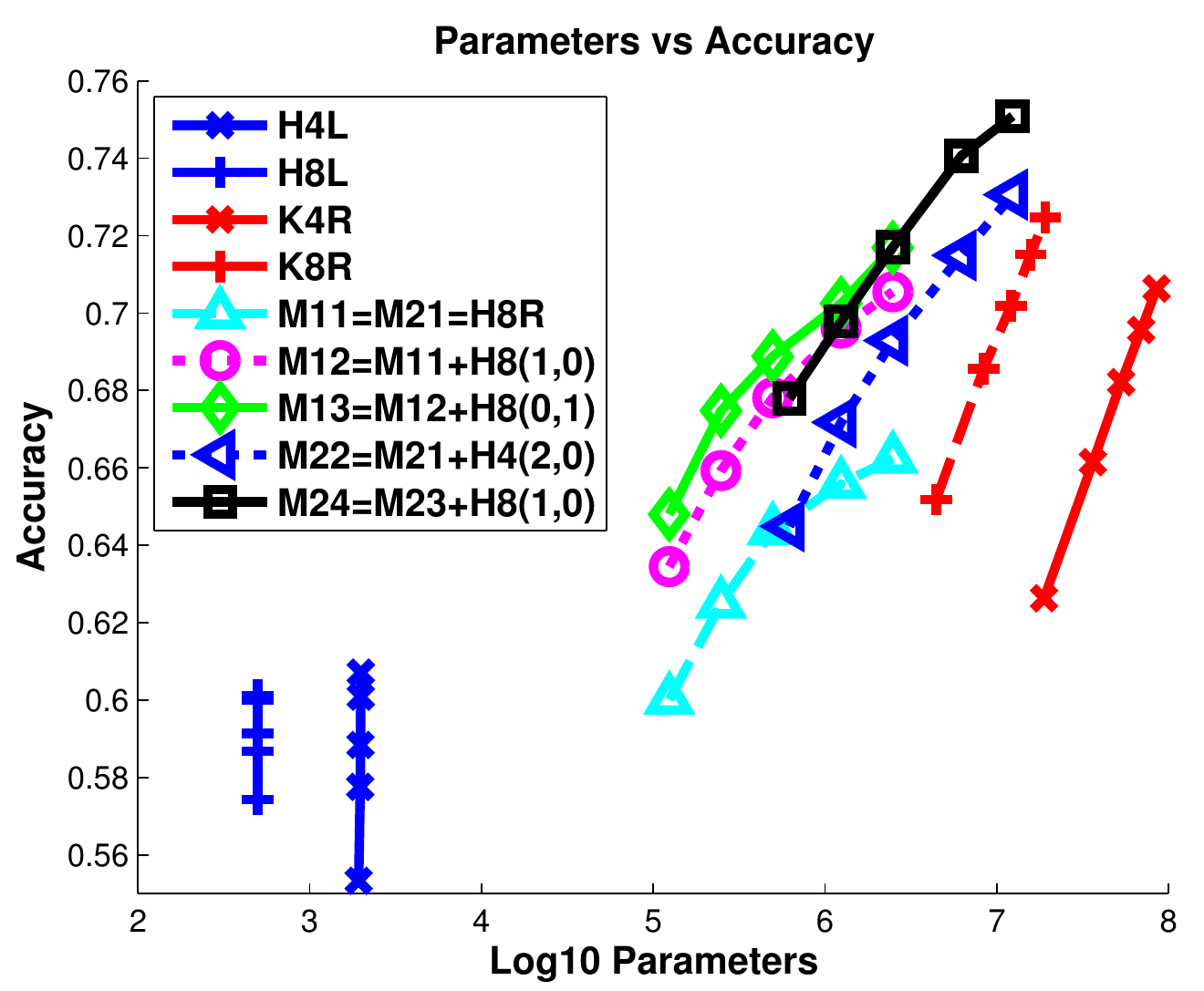}
		\caption{Parameters vs Performance}
		\label{paperE:fig:multiple_sims_params_support:params}
	\end{subfigure}
	\begin{subfigure}[b]{0.49\textwidth}
		\includegraphics[width=\textwidth]{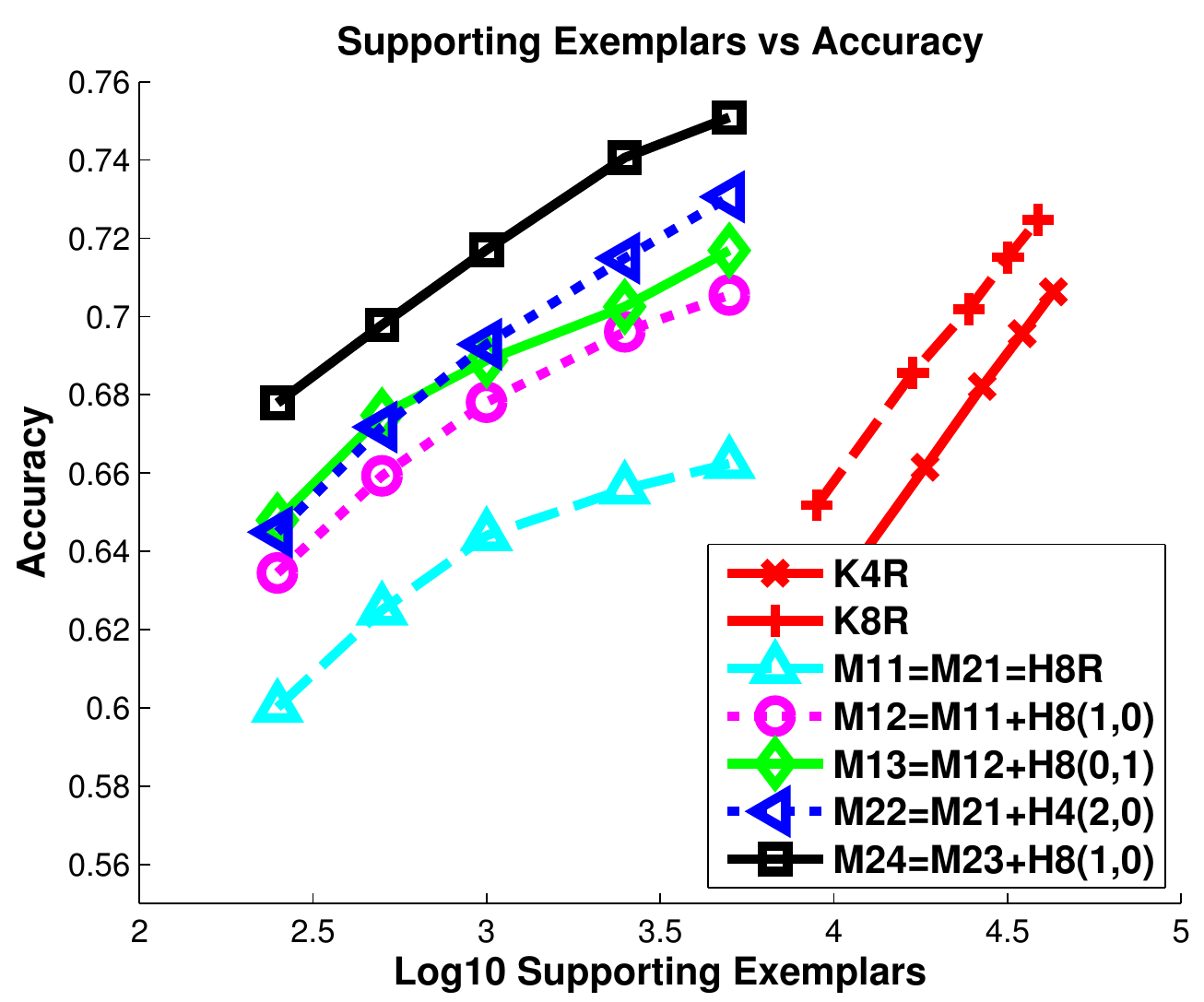}
		\caption{Supporting Exemplars vs Performance}
		\label{paperE:fig:multiple_sims_params_support:support}
	\end{subfigure}
\caption{Performance of BE-SVM vs model parameters for various sizes of the basis set, using multiple similarity measures.
Each curve for linear SVM (H4L, H8L) and kernelized SVM (K4R, K8R) represents the result for training on $1,\ldots,5$ folds of training data.
Each curve for BE-SVM shows the result for training model with a basis set of size $B=10\times\{25,50,100,250,500\}$ when trained on 5 folds of the training data.
} 
\label{paperE:fig:multiple_sims_params_support}
\end{figure}

We then trained BE-SVM models using these similarity measures for various sizes of the basis set, and for various sizes of training data.
Figures \ref{paperE:fig:multiple_sims_params_support:params} and \ref{paperE:fig:multiple_sims_params_support:support} show these results, where the BE-SVM models are trained on all 5 folds.
The shown number of supporting exemplars (and consequently the number of parameters) for BE-SVM are based on the size of the basis set.
It can be seen that using a basis size of $B=10\times 250$, the performance of the BE-SVM using more than 3 two-resolution similarity measures surpass that of the kernelized SVM trained on all the data and based on approximately $B=10\times4000$ support vectors.
Using low-resolution similarity measures, $B=10\times 500$ outperforms kernelized SVMs trained on up to 4 folds of the training data.
Furthermore, it can be observed that for the same model complexity, as measured either by the number of supporting exemplars, or by model parametrs, BE-SVM performs better than kernelized SVM.

\begin{figure}
\centering
\includegraphics[width=.66\linewidth]{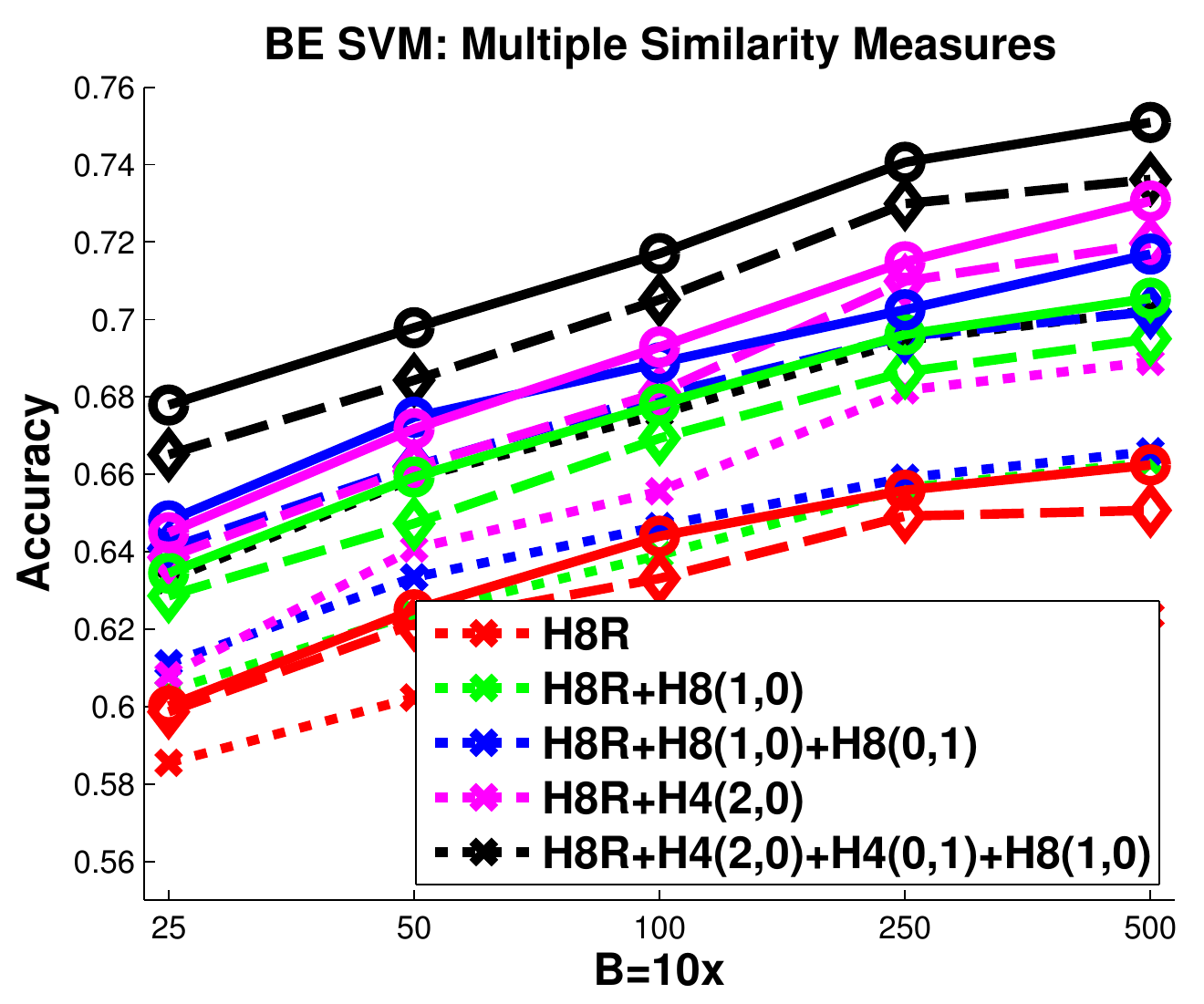}
\caption{Performance of BE-SVM using multiple similarity measures for various sizes of the basis set. 
Results with dotted, dashed, and solid lines represent 1, 3, and 5 folds worth of training data.
See text for analysis.
} 
\label{paperE:SM:fig:multiple_sims}
\end{figure}
Figure \ref{paperE:SM:fig:multiple_sims} shows the performance of BE-SVM using different similarity measures for various basis sizes and for different training set sizes.
It can be observed that using (invariant) indefinite similarity measures can significantly increase the performance of the model: compare the red curve with any other curve with the same line style.
For example, using all the training data and a two resolution deformable approach results in 8-10\% improvements in accuracy in comparison to the best performing PSD kernel (H8R).
Furthermore, the two-resolution approach outperforms the single resolution approach by approximately 3-4\% accuracy (compare blue and black curves with the same line style).

Measured by model parameters, BE-SVM is roughly 8 times sparser than kernelized SVM for the same accuracy.
Measured by supporting exemplars, its sparsity increases roughly to 30.
We need to point out that different similarity measures have different complexities \eg H8(1,0) is more expensive to evaluate than K8R. 
However, when the bases are shared for different similarity measures, CPU cache can be utilized much more efficiently as there will be less memory access and more (cached) computations.

\subsubsection{Multiple Kernel Learning with PSD Kernels}
We tried Multiple Kernel Learning (MKL) for kernelized SVM with PSD kernels.
When compared to sophisticated MKL methods, we found the following procedure to give competitive performances, with much less training costs.
Defining $K_C(.,.) = \alpha K_1(.,.) + (1-\alpha) K_2(.,.)$, our MKL approach consists of performing a line search for an optimal alpha $\alpha\in\left\{0,.1,\ldots,1\right\}$ which results in best 5-fold cross validating performance. 
Using this procedure, linear kernels were found not to contribute anything to Gaussian RBF kernels.
The optimal combination for high resolution and low resolution Gaussian RBF kernels (K4R and K8R) resulted in a performance gain of less than 0.5\% accuracy in comparison to K8R.
We founds this insignificant, and did not report its performance, considering the fact that the number of parameters increases approximately 4 times using this approach.

\subsection{BE-SVM's Normalizations}
It can be verified that in case of unnormalized features $\tilde\varphi^{(m)}(.)$, the corresponding Gram matrix will be 
\begin{equation}
\begin{array}{ll}
\tilde K(\mathbf{x}_i,\mathbf{x}_j) 	
				&= \sum_{m=1}^M\, \tilde K^{(m)}(\mathbf{x}_i,\mathbf{x}_j) 
				= \sum_{m=1}^M\, \sum_{b=1}^{B} \, s^{(m)}(\mathbf{b}_b,\mathbf{x}_i) s^{(m)}(\mathbf{b}_b,\mathbf{x}_j)
\end{array}
\label{paperE:eqn:besvm_mult_kernel}
\end{equation}
where $\tilde K^{(m)}$s are combined with equal weights, the value of each of which depends (locally) on how the similarities of $\mathbf{x}_i$ and $\mathbf{x}_j$ correlate with respect to the bases.
In the case of normalized features, the centered values of each similarity measure is weighted by $\left( \mathbb{E}_\mathcal{X}[\|\tilde\varphi - \mathbb{E}_\mathcal{X}[\tilde\varphi] \|] \right)^{-2}$ \ie more global weight is put on (the centered values of) the similarity measures with smaller variances in similarity values.

While the BE-SVM's normalization of empirical kernel maps is not optimal for discrimination, it can be seen as a reasonable prior for combining different similarity measures.
Utilizing such a prior, in combination with linear classifiers and $\ell_P$ regularizers, has two important consequences: 1) the centering helps reduce the correlation between dimensions and the scaling helps balance the effect of regularization on different similarity measures, irrespective of their overall norms, and 2) such a scaling directly affects the parameter tuning for learning the linear classifiers: for all the similarity measures (and combinations of similarity measures) with various basis sizes, the same parameter: $C=1$ was used to train the classifiers. 
While cross-validation will still be a better option, cross-validating for different parameters settings -- and specially when combining multiple similarity measures -- will be very expensive and prohibitive. 
By using the BE-SVM's normalization, we essentially avoid searching for optimal combining weights for different similarity measures and also tuning for the $C$ parameter of the linear SVM training.

In this section, we quantitatively evaluate the normalization suggested for BE-SVM (\ref{paperE:eqn:besvm_feature_map_norm}), and compare it to a few other combinations.
Particularly, we consider various normalizations of the HOG feature vectors, and similarly, various normalization schemes for the empirical kernel map $\tilde{\varphi}$ (\ref{paperE:eqn:besvm_feature_map}).
We consider the following normalizations:
\begin{itemize}
\item No normalization (Unnorm)
\item Z-Scoring, namely centering and scaling each dimension by the inverse of its standard deviation (Z-Score)
\item BE-SVM normalization, namely centering and scaling all dimensions by the inverse average $\ell_2$ norm of the centered vectors (BE-SVM)
\end{itemize}
We report test performances for all combinations of normalizations for the feature vectors and the empirical kernel maps, for two cases: 1) when $C=1$, and 2) when the $C$ parameter is cross-validated from $\mathcal{C}=\{10^{-1}, 10^0, 10^1\}$.
In both cases, $|\mathcal{B}| = 10 \times 100$ bases were uniformly sub-sampled from the first fold of the training set (`Indx' basis selection).

\begin{figure}
\centering
\includegraphics[width=.49\linewidth]{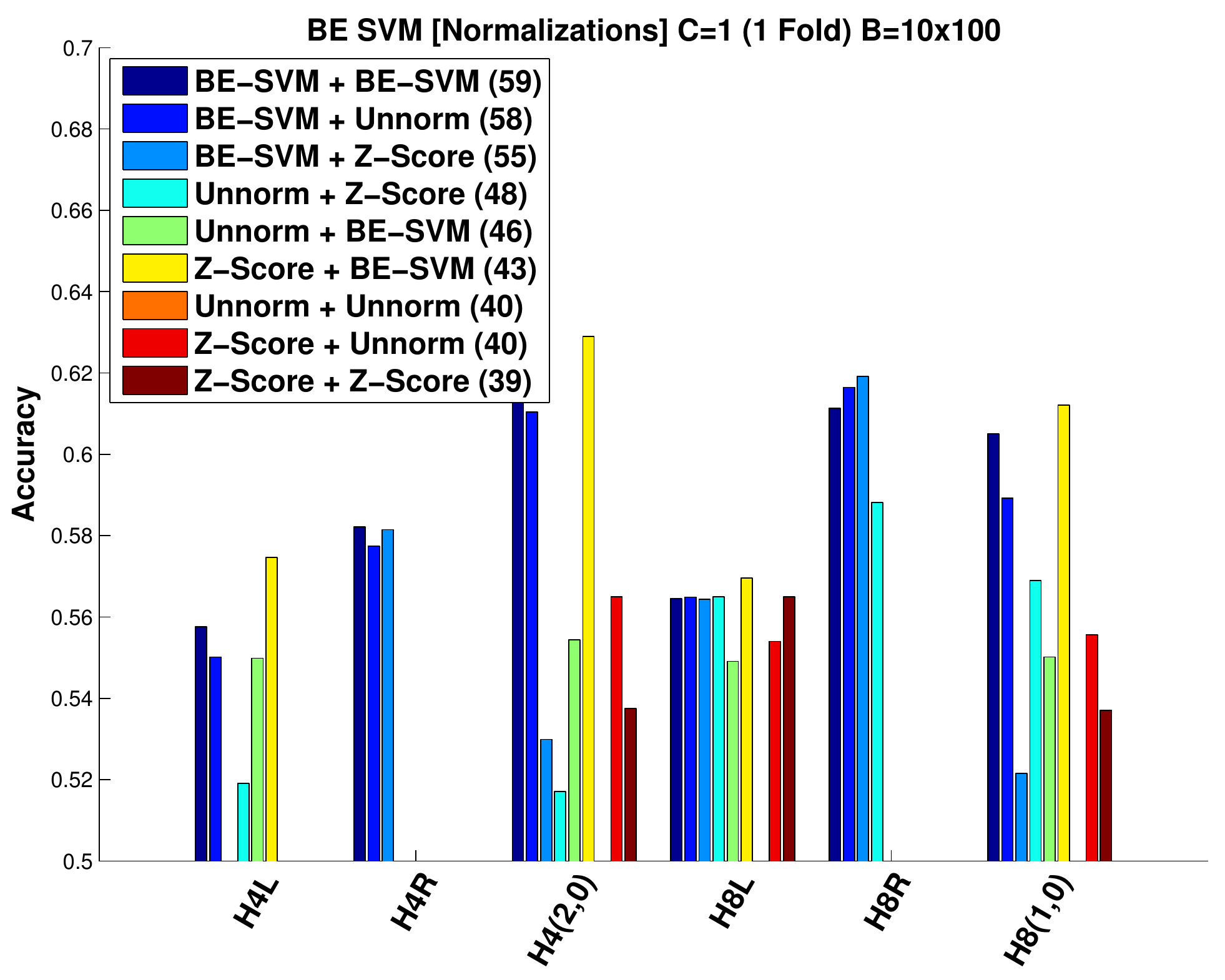}
\includegraphics[width=.49\linewidth]{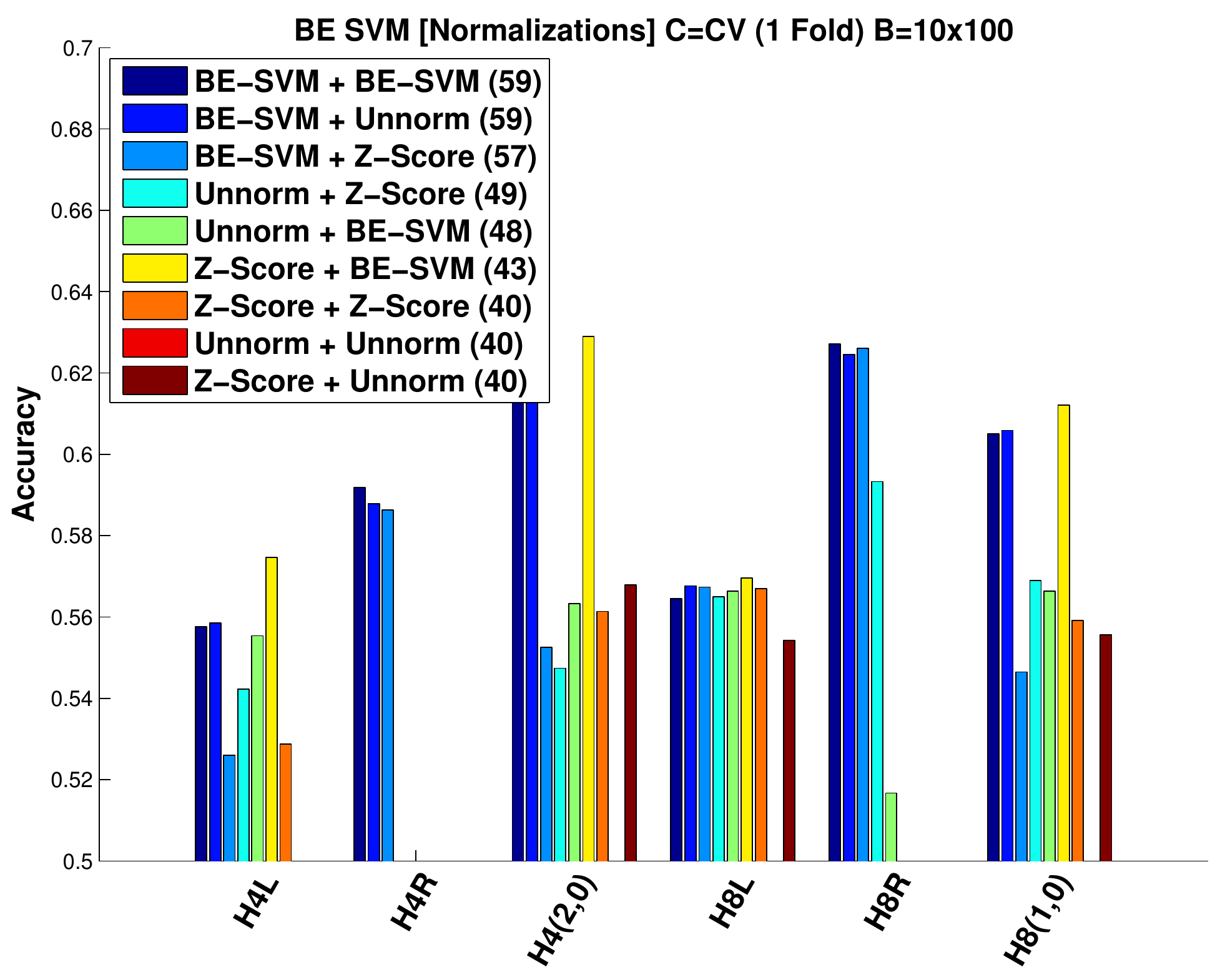}
\caption{Performance of BE-SVM for different normalization schemes of the feature vector and the empirical kernel map, and different similarity measures. ``F + K (P)'' in the legend reflects using F and K normalization schemes for the feature vectors and the empirical kernel maps respectively, which results in the average test performance of P (averaged over the similarity measures).} 
\label{paperE:SM:fig:normalization_one_sim}
\end{figure}

\begin{figure}
\centering
\includegraphics[width=.49\linewidth]{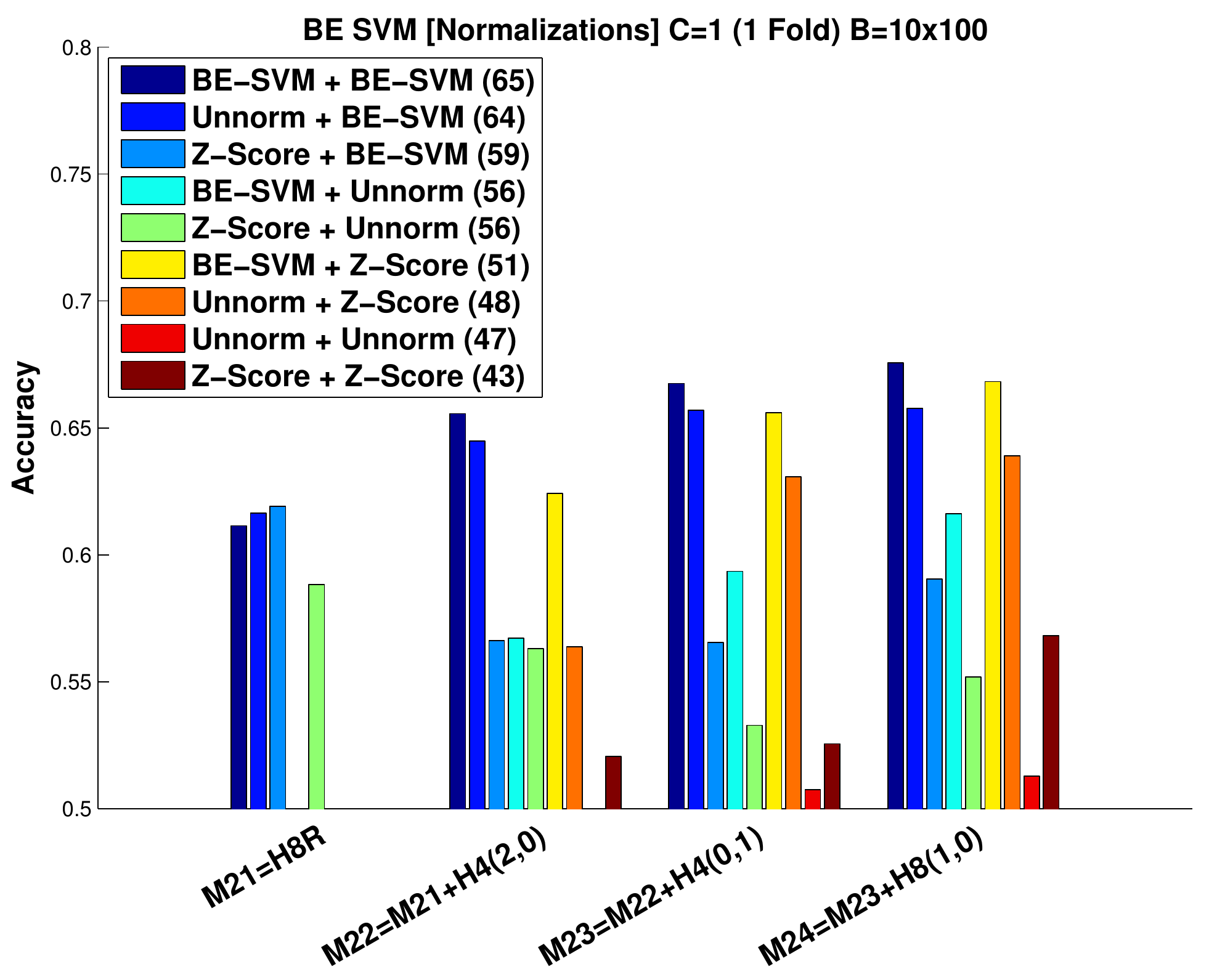}
\includegraphics[width=.49\linewidth]{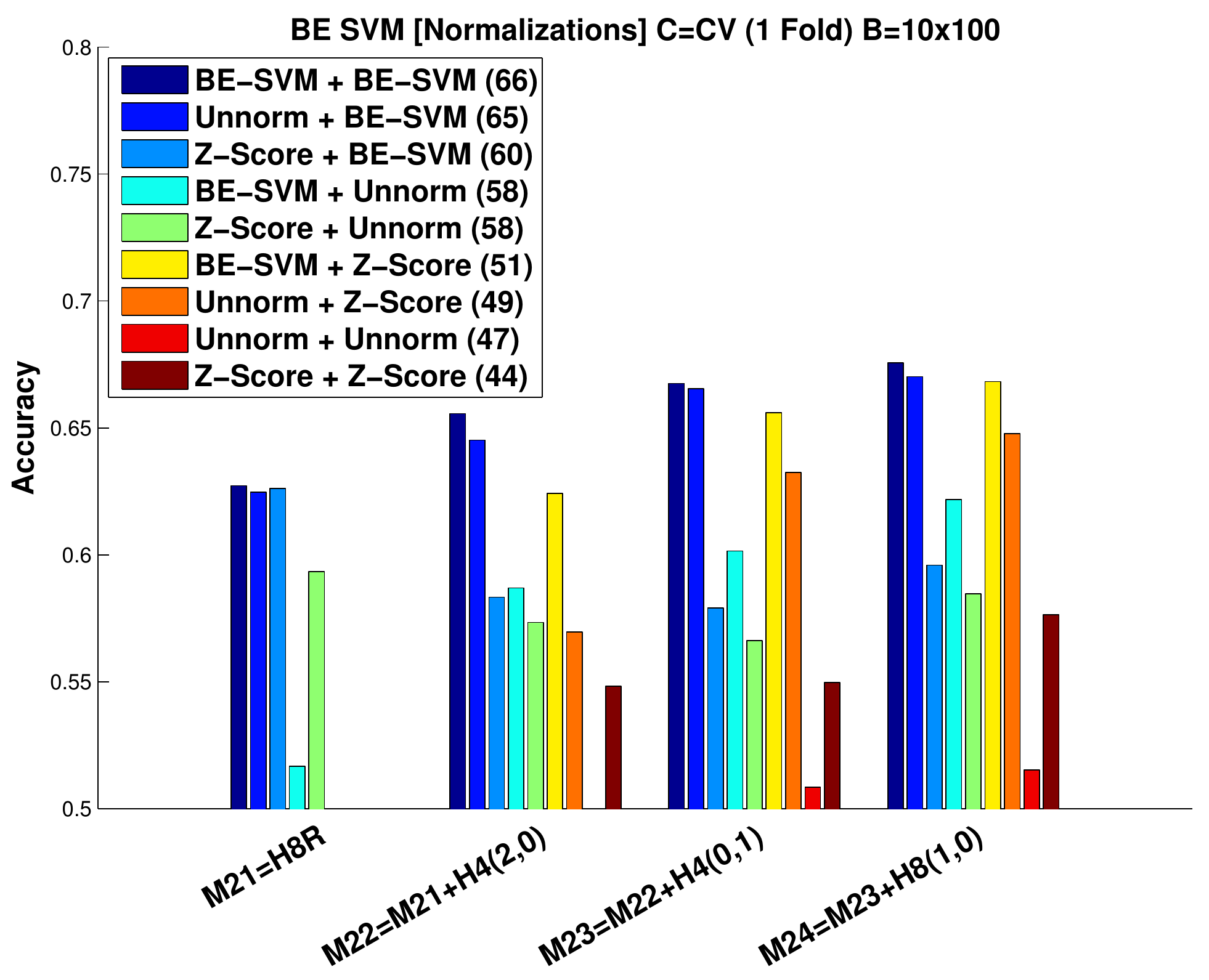}
\caption{Performance of BE-SVM for different normalization schemes of the feature vector and the empirical kernel map, and different combinations of similarity measures. ``F + K (P)'' in the legend reflects using F and K normalization schemes for the feature vectors and the empirical kernel maps respectively, which results in the average test performance of P (averaged over the combinations of similarity measures).} 
\label{paperE:SM:fig:normalization_multi_sim}
\end{figure}

Figure \ref{paperE:SM:fig:normalization_one_sim} shows the performance of BE-SVM in combination with different normalizations of the feature vectors and empirical kernel maps, and for different similarity measures. 
On top, reported numbers are for $C=1$ while on the bottom, $C$ is cross validated. 
It can be observed that the BE-SVM's normalization works best both for the feature and empirical kernel map normalizations.
Although z-scoring is more suitable for linear similarity measures (compare BE-SVM + BE-SVM with Z-SCORE + BE-SVM in H4L, H8L, H4(x,y) and H8(x,y)), overall BE-SVM's normalization of the feature space works better than the alternatives.
Particularly, in single similarity measure cases, it seems that normalizing the feature according to the BE-SVM's normalization is more important than normalizing the empirical kernel map.
While the cross-validation of the $C$ parameter marginally affects the performance, it does not change the conclusions drawn from the $C=1$ case.

Figure \ref{paperE:SM:fig:normalization_multi_sim} shows the performance of BE-SVM in combination with different normalizations of the feature vectors and empirical kernel maps, and for different combinations of similarity measures (the sequence of greedily augmented similarity measures $\mathcal{M}_2$: the set of two resolution similarity measures described in Section \ref{paperE:section:multiple_sims}). 
It can be observed that BE-SVM's normalization of the kernel map is much more important and effective when combining multiple similarity measure (compare to Figure \ref{paperE:SM:fig:normalization_one_sim}) .

These observations quantitatively motivate the use of BE-SVM's normalization with the following benefits, at least on the dataset we experimented on:
\begin{itemize}
\item It removes the need for cross-validation for tuning the $C$ parameter, and mixing weights for different similarity measures.
\item As the feature vector is centered and properly scaled, the linear SVM solver converges much faster than the unnormalized case, or when $C>>1$.
\item It results in robust learning of BE-SVM which can efficiently combine different similarity measures \ie RBF kernels (H8R), and linear deformable similarity measures (H4(2,0), H4(0,1), H8(1,0)).
\end{itemize}

\section{Conclusion}
\label{paperE:sec:conclusion}
We analyzed scalable approaches for using indefinite similarity measures in large margin scenarios.
We showed that our model based on an explicit basis expansion of the data according to arbitrary similarity measures can result in competitive recognition performances, while scaling better with respect to the size of the data.
The model named Basis Expanding SVM was thoroughly analyzed and extensively tested on CIFAR-10 dataset.

In this study, we did not explore basis selection strategies, mainly due to the small intra-class variation of the dataset.
We expect basis selection strategies to play a crucial role in the performance of the resulting model on more challenging datasets \eg Pascal VOC or ImageNet.
Therefore, an immediate future work is to apply BE-SVM to larger scale and more challenging problems \eg object detection, in combination with data driven basis selection strategies.

{\small
\bibliographystyle{ieee}
\bibliography{arXiv}

\begin{thebibliography}{10}\itemsep=-1pt

\bibitem{AghazadehASC12}
O.~Aghazadeh, H.~Azizpour, J.~Sullivan, and S.~Carlsson.
\newblock Mixture component identification and learning for visual recognition.
\newblock In {\em European Conference on Computer Vision}, 2012.

\bibitem{CC01a}
C.-C. Chang and C.-J. Lin.
\newblock {LIBSVM}: A library for support vector machines.
\newblock {\em ACM Transactions on Intelligent Systems and Technology}, pages
  27:1--27:27, 2011.

\bibitem{sim_class}
Y.~Chen, E.~K. Garcia, M.~R. Gupta, A.~Rahimi, and L.~Cazzanti.
\newblock Similarity-based classification: Concepts and algorithms.
\newblock {\em Journal of Machine Learning Research}, pages 747--776, 2009.

\bibitem{learning_kernels_from_indef}
Y.~Chen, M.~R. Gupta, and B.~Recht.
\newblock Learning kernels from indefinite similarities.
\newblock In {\em International Conference on Machine Learning}, 2009.

\bibitem{DT05}
N.~Dalal and B.~Triggs.
\newblock Histograms of oriented gradients for human detection.
\newblock In {\em IEEE Conference on Computer Vision and Pattern Recognition},
  2005.

\bibitem{invariant_kernels}
D.~Decoste and B.~Sch\"{o}lkopf.
\newblock Training invariant support vector machines.
\newblock {\em Machine Learning}, pages 161--190, 2002.

\bibitem{graph_matching}
O.~Duchenne, A.~Joulin, and J.~Ponce.
\newblock {A Graph-matching Kernel for Object Categorization}.
\newblock In {\em IEEE International Conference on Computer Vision}, 2011.

\bibitem{REF08a}
R.-E. Fan, K.-W. Chang, C.-J. Hsieh, X.-R. Wang, and C.-J. Lin.
\newblock {LIBLINEAR}: A library for large linear classification.
\newblock {\em Journal of Machine Learning Research}, 2008.

\bibitem{FelzenszwalbGMR10}
P.~F. Felzenszwalb, R.~B. Girshick, D.~A. McAllester, and D.~Ramanan.
\newblock Object detection with discriminatively trained part-based models.
\newblock {\em IEEE Transactions on Pattern Analysis and Machine Intelligence},
  pages 1627--1645, 2010.

\bibitem{mixing_svm}
Z.~Fu, A.~Robles-Kelly, and J.~Zhou.
\newblock Mixing linear svms for nonlinear classification.
\newblock {\em Neural Networks}, pages 1963--1975, 2010.

\bibitem{pairwise_proximity}
T.~Graepel, R.~Herbrich, P.~Bollmann-Sdorra, and K.~Obermayer.
\newblock Classification on pairwise proximity data.
\newblock In {\em Neural Information Processing Systems}, 1998.

\bibitem{classification_proximity_lp}
T.~Graepel, R.~Herbrich, B.~Sch\"{o}lkopf, A.~Smola, P.~Bartlett,
  K.~M\"{u}ller, K.~Obermayer, and R.~Williamson.
\newblock Classification on proximity data with lp-machines.
\newblock In {\em Neural Information Processing Systems}, 1999.

\bibitem{Indefinite_Kernels}
B.~Haasdonk.
\newblock Feature space interpretation of svms with indefinite kernels.
\newblock {\em IEEE Transactions on Pattern Analysis and Machine Intelligence},
  2005.

\bibitem{Lee01rsvm:reduced}
Y.~jye Lee and O.~L. Mangasarian.
\newblock Rsvm: Reduced support vector machines.
\newblock In {\em Data Mining Institute, Computer Sciences Department,
  University of Wisconsin}, 2001.

\bibitem{Keerthi06buildingsupport}
S.~S. Keerthi, O.~Chapelle, and D.~DeCoste.
\newblock Building support vector machines with reduced classifier complexity.
\newblock {\em Journal of Machine Learning Research}, pages 1493--1515, 2006.

\bibitem{deformable_models}
D.~Keysers, T.~Deselaers, C.~Gollan, and H.~Ney.
\newblock Deformation models for image recognition.
\newblock {\em IEEE Transactions on Pattern Analysis and Machine Intelligence},
  pages 1422--1435, 2007.

\bibitem{Krizhevsky09learningmultiple}
A.~Krizhevsky.
\newblock Learning multiple layers of features from tiny images.
\newblock Technical report, 2009.

\bibitem{similarity_svm}
L.~Liao and W.~S. Noble.
\newblock {Combining Pairwise Sequence Similarity and Support Vector Machines
  for Detecting Remote Protein Evolutionary and Structural Relationships}.
\newblock {\em Journal of Computational Biology}, pages 857--868, 2003.

\bibitem{ong04leanpk}
C.~S. Ong, X.~Mary, S.~Canu, and A.~J. Smola.
\newblock Learning with non-positive kernels.
\newblock In {\em International Conference on Machine Learning}, 2004.

\bibitem{defense_onevall}
R.~Rifkin and A.~Klautau.
\newblock In defense of one-vs-all classification.
\newblock {\em Journal of Machine Learning Research}, pages 101--141, 2004.

\bibitem{good_rec_nonmetric}
W.~J. Scheirer, M.~J. Wilber, M.~Eckmann, and T.~E. Boult.
\newblock Good recognition is non-metric.
\newblock {\em CoRR}, 2013.

\bibitem{Steinwart04sparsenessof}
I.~Steinwart.
\newblock Sparseness of support vector machines -- some asymptotically sharp
  bounds.
\newblock In {\em Neural Information Processing Systems}, 2004.

\bibitem{Williams00theeffect}
C.~Williams and M.~Seeger.
\newblock The effect of the input density distribution on kernel-based
  classifiers.
\newblock In {\em International Conference on Machine Learning}, 2000.

\bibitem{direct_sparse}
M.~Wu, B.~Sch\"{o}lkopf, and G.~Bakir.
\newblock A direct method for building sparse kernel learning algorithms.
\newblock {\em Journal of Machine Learning Research}, pages 603--624, 2006.

\bibitem{YangWVM12}
W.~Yang, Y.~Wang, A.~Vahdat, and G.~Mori.
\newblock Kernel latent svm for visual recognition.
\newblock In {\em Neural Information Processing Systems}, 2012.

\end{thebibliography}
}

\end{document}